%% file: main.tex
\documentclass[nohyperref]{article}

\usepackage{microtype}
\usepackage{graphicx}
\usepackage{booktabs} 
\usepackage{multicol}
\usepackage{multirow}
\usepackage{floatpag}
\usepackage{wrapfig, floatrow}
\usepackage[font=small]{caption}
\usepackage{float}
\usepackage[round,authoryear]{natbib}
\usepackage{afterpage}
\usepackage{listings}
\usepackage{hyperref}
\usepackage[normalem]{ulem} 
\usepackage{hyperref}
\usepackage{color}
\definecolor{codegreen}{rgb}{0,0.5,0}
\definecolor{codeblue}{rgb}{0,0,0.9}
\definecolor{codeblues}{rgb}{0,0,0.4}
\definecolor{codegray2}{rgb}{0.4,0.4,0.4}
\definecolor{codegray}{rgb}{0.9,0.9,0.9}
\definecolor{codepurple}{rgb}{0.58,0,0.82}
\definecolor{backcolour}{rgb}{0.95,0.95,0.92}
\definecolor{backcolour2}{rgb}{0.9,0.9,0.9}
\definecolor{codered}{rgb}{0.5,0,0}
\definecolor{textcodered}{rgb}{0.05,0.05,0.05}
\definecolor{palegray}{rgb}{0.98,0.98,0.99}
\usepackage{textcomp}

\usepackage{amssymb}
\usepackage{amsthm}
\usepackage{amsmath}
\usepackage[normalem]{ulem} 

\usepackage{xspace}
\usepackage{colortbl}  
\usepackage{subcaption}
\usepackage{hhline}
\usepackage{pifont}
\usepackage{threeparttable}
\usepackage{makecell}
\input{math.tex}

\newcommand{\eg}{\emph{e.g.}}
\newcommand{\ie}{\emph{i.e.}}

\newcommand{\model}{CtrlFormer\xspace}

\newcommand{\onepm}[2]{\text{#1}$_{\pm \text{#2}}$}
\newcommand{\onebfpm}[2]{\textbf{#1}$_{\pm \text{#2}}$}

\definecolor{walkercolor}{RGB}{244,157,78}
\newcommand{\walkercolor}[1]{\textcolor{walkercolor}{#1}}
\definecolor{reachercolor}{RGB}{103, 78, 167}
\newcommand{\reachercolor}[1]{\textcolor{reachercolor}{#1}}
\definecolor{cartpolecolor}{RGB}{153, 51, 102}
\newcommand{\cartpolecolor}[1]{\textcolor{cartpolecolor}{#1}}
\definecolor{fingercolor}{RGB}{80, 200, 180}
\newcommand{\fingercolor}[1]{\textcolor{fingercolor}{#1}}

\usepackage[accepted]{icml2022}

\usepackage{amsmath}
\usepackage{amssymb}
\usepackage{mathtools}
\usepackage{amsthm}

\usepackage[capitalize,noabbrev]{cleveref}

\theoremstyle{plain}

\theoremstyle{definition}

\theoremstyle{remark}

\usepackage[textsize=tiny]{todonotes}

\icmltitlerunning{\model: Learning Transferable State Representation for 
Visual Control via Transformer}

\begin{document}

\twocolumn[
\icmltitle{\model: Learning Transferable State Representation for \\
Visual Control via Transformer}

\begin{icmlauthorlist}
\icmlauthor{Yao Mu}{yyy}
\icmlauthor{Shoufa Chen}{yyy}
\icmlauthor{Mingyu Ding}{yyy}
\icmlauthor{Jianyu Chen}{tsing}
\icmlauthor{Runjian Chen}{yyy}
\icmlauthor{Ping Luo}{yyy}
\end{icmlauthorlist}

\icmlaffiliation{yyy}{Department of Computer Science, the University of Hong Kong, Hong Kong}
\icmlaffiliation{tsing}{Institute for Interdisciplinary Information Sciences (IIIS), Tsinghua University, Beijing, China}

\icmlcorrespondingauthor{Ping Luo}{pluo@cs.hku.hk}

\icmlkeywords{Machine Learning, ICML}

\vskip 0.3in
]

\printAffiliationsAndNotice{}  

\begin{abstract}
Transformer has achieved great successes in learning vision and language representation, which is general across various downstream tasks. In visual control, learning transferable state representation that can transfer between different control tasks is important to reduce the training sample size. However, porting Transformer to sample-efficient visual control remains a challenging and unsolved problem.
To this end, we propose a novel  Control Transformer (CtrlFormer), possessing many appealing benefits that prior arts do not have.
Firstly, CtrlFormer jointly learns self-attention mechanisms between visual tokens and policy tokens among different control tasks, where multitask representation can be learned and transferred without catastrophic forgetting.
Secondly, we carefully design a contrastive reinforcement learning paradigm to train CtrlFormer, enabling it to achieve high sample efficiency, which is important in control problems. 
For example, in the DMControl benchmark, unlike recent advanced methods that failed by producing a zero score in the ``Cartpole'' task after transfer learning with 100$k$ samples, CtrlFormer can achieve a state-of-the-art score \texttt{$769_{\pm{34}}$} with only 100$k$ samples, while maintaining the performance of previous tasks.
The code and models are released in our \href{https://sites.google.com/view/ctrlformer-icml/}{project homepage}.

\end{abstract}

\input{src/1-intro}

\input{src/5-survey}

\input{src/2-settings}

\input{src/3-method}

\input{src/4-exp}

\input{src/6-conclusion}

\bibliography{main}
\bibliographystyle{icml2022}
\clearpage
\input{src/7-appendix}

\end{document}

%% file: math.tex
\usepackage{amsmath,amsfonts,bm,amssymb}

\def\eqref#1{equation~\ref{#1}}

\def\1{\bm{1}}

\DeclareMathAlphabet{\mathsfit}{\encodingdefault}{\sfdefault}{m}{sl}
\SetMathAlphabet{\mathsfit}{bold}{\encodingdefault}{\sfdefault}{bx}{n}



%% file: src/1-intro.tex
\newcommand{\netparams}{{\theta}}
\newcommand{\targetparams}{\xi}
\newcommand{\augview}{augmented view}
\newcommand{\aaugview}{an augmented view}
\newcommand{\imageaugmentation}{image augmentation}
\newcommand{\Imageaugmentation}{Image augmentation}
\newcommand{\augmentation}{augmentation}
\newcommand{\mono}[1]{\texttt{\color{textcodered}#1}}
\newcommand{\dono}[1]{\texttt{\color{codeblues}#1}}
\newcommand{\myfigure}[2]{
\begin{figure}[H]
\floatbox[{\capbeside\thisfloatsetup{capbesideposition={right,top},
capbesidewidth=0.65\textwidth}}]{figure}[60pt]
{\caption{#1}}
{\vspace{10pt}\includegraphics[width=2.5cm]{#2}}
\end{figure}
\vspace{-.5cm}
}

\section{Introduction}
\label{submission}

\begin{figure}[ht]
\footnotesize
\begin{center}
\begin{subfigure}{0.49\textwidth}
\includegraphics[width=0.99\textwidth]{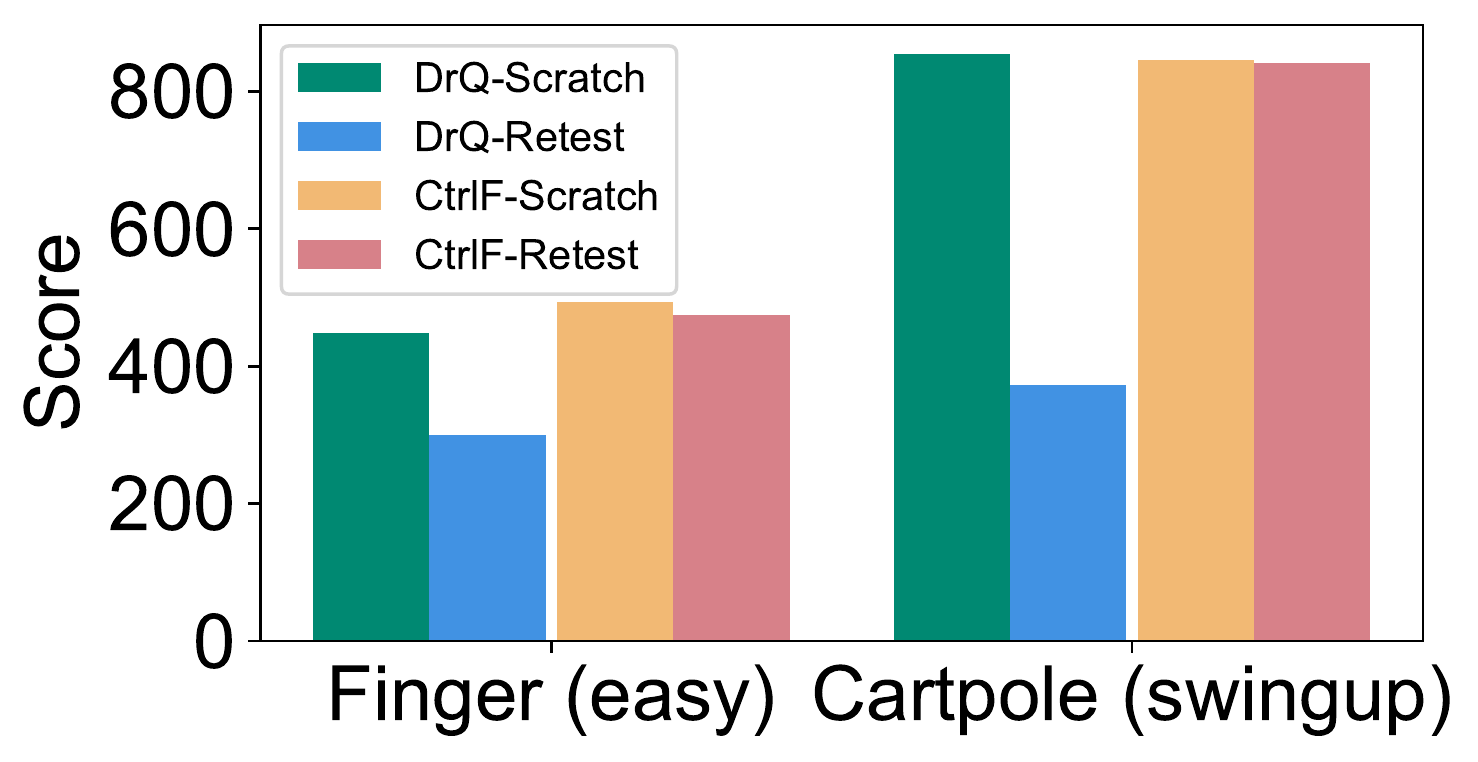}
\caption{Maintainability}\label{fig:maintain}
\end{subfigure}
\begin{subfigure}{0.49\textwidth}
\includegraphics[width=0.99\textwidth]{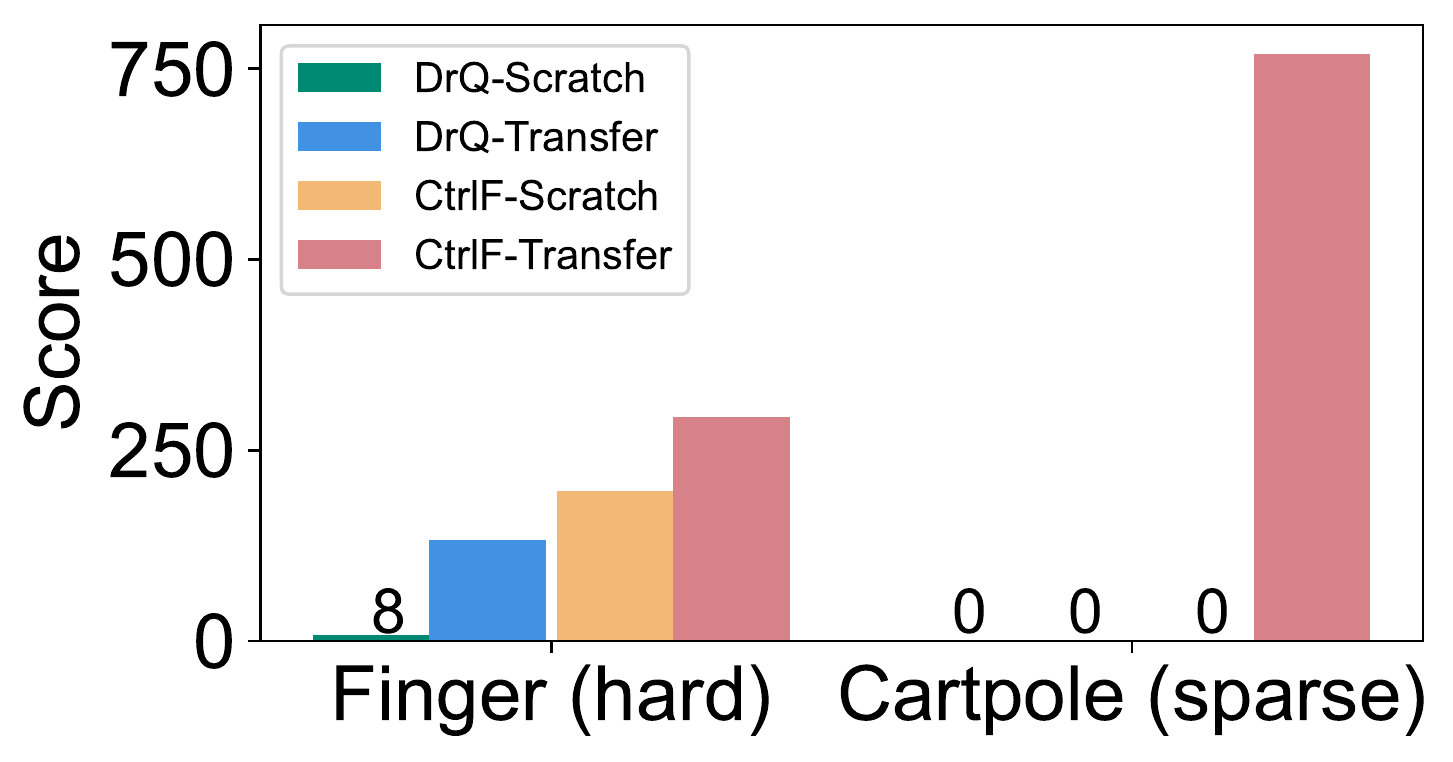}
\caption{Transferability}\label{fig:trans}
\end{subfigure}
\end{center}
\vspace{-10pt}
\caption{\textbf{Effect of \model.} The agent first learns state representations in one task, and then transfers them to a new task, \eg, from finger~(turn-easy) to finger~(turn-hard), and from cartpole~(swingup) to cartpole~(swingup-sparse). Figure~\textbf{\ref{fig:maintain}} shows the \textit{maintainability} by comparing the performance in the old task before~(scratch) and after~(retest) transferring to a new task. \model doesn't have catastrophic forgetting after transferring to the new task, and the performance is basically the same as learning from scratch. However, the performance of DrQ~\cite{kostrikov2020image} drops significantly after transferring; Figure~\textbf{\ref{fig:trans}} shows the \textit{transferability} by comparing the performance of learning from scratch and from transferring.
Transferring previous learned knowledge benefits much for \model~(labeled as CtrlF).}\label{fig:Effect}\vspace{-5pt}
\end{figure}

Visual control is important for various real-world applications such as playing Atari games~\cite{mnih2015human}, Go games~\cite{silver2017mastering}, robotic control~\cite{kober2013reinforcement,yuan2022don}, and autonomous driving~\cite{wang2018deep}.
Although remarkable successes have been made, a long-standing goal of visual control, transferring the learned knowledge to new tasks without catastrophic forgetting, remains challenging and unsolved.

Unlike a machine, a  human can quickly identify the critical information to learn a new task with only a few actions and observations, since a human can discover the relevancy/irrelevancy between the current task and the previous tasks he/she has learned, and decides which information to keep or transfer.
As a result, a new task can be learned quickly by a human without forgetting what has been learned before. Moreover, the ``state representation'' can be strengthened for better generalization to future tasks.

Modern machine learning methods for transferable representation learning across tasks can be generally categorized into three streams. They have certain limitations.
In the first stream, many representative works such as~\cite{shah2021rrl} utilize features pretrained in the ImageNet~\cite{deng2009imagenet} and COCO~\cite{lin2014microsoft} datasets.
The domain gap between the pretraining datasets and the target datasets hinders their performance and sample efficiency.
In the second stream, \citet{rusu2016progressive, fernando2017pathnet} trained a super network 
to accommodate a new task. These approaches often allocate different parts of the supernet to learn different tasks. However, the network parameters and computations are proportionally increased when the number of tasks increases.
In the third stream, the latent variable models~\cite{ha2018world,hafner2019learning,hafner2019dream}  learn representation by optimizing the variational bound. These approaches are struggling when the tasks come from different domains.

Can we learn sample-efficient transferable state representation across different control tasks in a single Transformer network?
The self-attention mechanism in Transformer mimics the perceived attention of synaptic connections in the human brain~as argued in \cite{oby2019new}.
Transformer could be powerful to model the relevancy/irrelevancy between different control tasks to alleviate the weaknesses in previous works.
However, simply porting Transformer to this problem cannot solve the above limitations because Transformer is extremely sample-inefficient (data-hungry) as demonstrated in  NLP~\cite{vaswani2017attention, devlin2018bert} and computer vision~\cite{dosovitskiy2020image}.

\begin{figure}[t]
    \centering
    \includegraphics[width=0.9\linewidth]{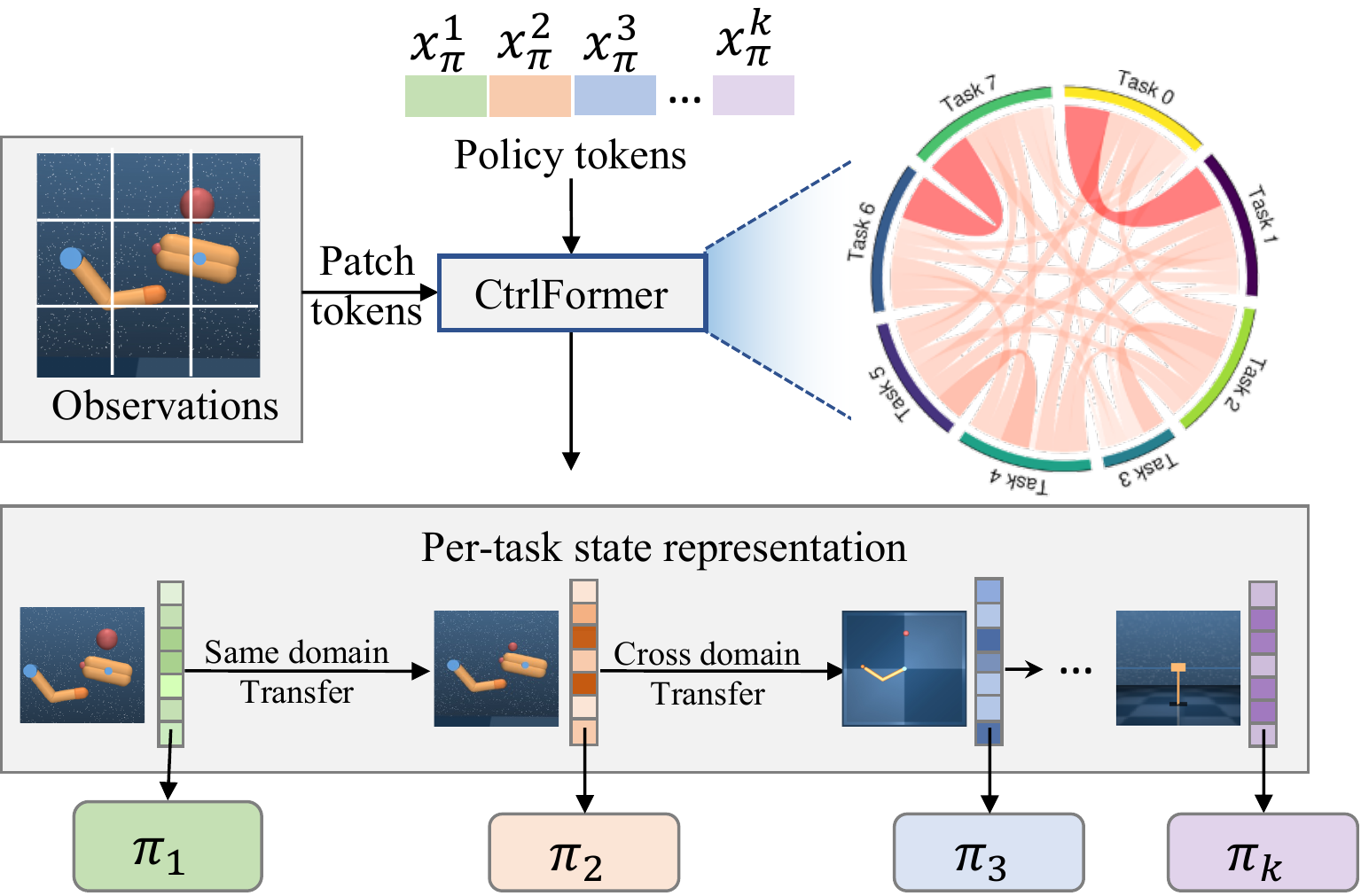}
    \vspace{-5pt}
    \caption{\footnotesize
    \textbf{Overview of \model for visual control.} 
    The input image is split into several patch tokens. Each task is assigned a specific policy token, which is a randomly-initialized and learnable variable. Tasks can come from the same domain, such as the Finger-turn-easy and Finger-turn-hard, or cross-domain, such as Reacher-easy and Cartpole-swingup. 
    \model learns the self-attention between observed image tokens, as well as policy tokens of
    different control tasks by vision transformer, which helps the agent leverage the similarities with previous tasks and reuse the representations from previously learned tasks to promote the behaviour learning of the current task. The output of \model is used as the input of the downstream policy networks. 
    }\vspace{-5pt}
\label{fig:Overview}
\end{figure}

This paper proposes a novel Transformer for representation learning in visual control, named \model, which has two benefits compared to previous works.
Firstly, the self-attention mechanism in \model learns both visual tokens and policy tokens for multiple different control tasks simultaneously, fully capturing relevant/irrelevant features between tasks.  This enables the knowledge learned from previous tasks can be transferred to a new task,  while maintaining the learned representation of previous tasks. 
Secondly, CtrlFormer reduces training sample size by using contrastive reinforcement learning, where the gradients of the policy loss  and the self-supervised contrastive loss are propagated jointly for representation learning.
For example,
as shown in Figure \ref{fig:Overview}, the input image is divided into several patches, and each patch corresponds to a token. The CtrlFormer learns self-attentions between image tokens, as well as policy tokens of different control tasks such as ``standing" and ``walking".
In this way, CtrlFormer not only decouples multiple control policies, but also decouples the features for behaviour learning and self-supervised visual representation learning, improving transferability among different tasks. 

Our contributions are three-fold. \textbf{(1)} A novel control Transformer (\model) is proposed for learning transferable state representation in visual control tasks. \model models the relevancy/irrelevancy between distinct tasks by self-attention mechanism across visual data and control policies. It makes the knowledge learned from previous tasks transferable to a new task, while maintaining accuracy and high sample efficiency in previous tasks.
\textbf{(2)} \model can improve sample efficiency by combining reinforcement learning with self-supervised contrastive visual representation learning~\cite{he2020momentum, grill2020bootstrap} and can reduce the number of parameters and computations via a pyramid Transformer structure.
\textbf{(3)} Extensive experiments show that \model outperforms previous works in terms of both transferability and sample efficiency. As shown in Figure~\ref{fig:Effect},
transferring previously learned state representation significantly improves the sample efficiency of learning new tasks. Furthermore, \model does not have catastrophic forgetting after transferring to the new task, and the performance is basically the same as learning from scratch.

%% file: src/5-survey.tex
\section{Related Works}\label{sec:related_works}

\textbf{Learning Transferable State Representation.} For task-specific representation learning tasks like classification and contrastive objectives, the representation learned on large-scale offline datasets~(ImageNet~\cite{deng2009imagenet}, COCO~\cite{lin2014microsoft}, and etc.) has high generalization ability~\cite{he2020momentum, yen2020learning}. However, the downstream reinforcement learning methods based on such a task-agnostic representation empirically show low sample efficiency since the representation contains a lot of task-irrelevant interference information.

Progressive neural network~\cite{rusu2016progressive, rusu2017sim,gideon2017progressive} is a representative structure of transferring state representations, which
is composed of multiple columns, where each column is a policy network for a specific task, and lateral connections are added. 
The parameters need to learn grow proportionally with the number of incoming tasks, which hinders its scalability.

PathNet~\cite{fernando2017pathnet} tries to alleviate this issue by using a size-fixed network. It contains multiple pathways, which are subsets of neurons whose weights contain the knowledge of previous tasks and are frozen during training on new tasks. The pathways that determine which parts of the network could be re-used for new tasks are discovered by a tournament selection genetic algorithm. PathNet fixes the parameters along a path learned on previously learned tasks and re-evolves a new population of paths for a new task to accelerate the behavior learning.  However, the process of pathways discovery with the genetic algorithm has a high cost on the computational resources.

Latent variable models offer a flexible way to represent key information of the observations by optimizing the variational lower bound~\cite{krishnan2015deep, karl2016deep, doerr2018probabilistic, buesing2018learning,ha2018world, hafner2019learning, hafner2019dream, tirinzoni2020sequential,mu2021modelbased,chen2022flow}. \citet{2018Recurrent} propose the world model algorithm to learn representation by variational autoencoder~(VAE). PlaNet~\cite{hafner2019learning} utilizes a recurrent stochastic state model~(RSSM) to learn the representation and latent dynamic jointly. The transition probability is modeled on the latent space instead of the original state space. 
Dreamer~\cite{hafner2019dream} utilizes the RSSM to make the long-term imagination. Although Dreamer is promising to transfer the knowledge across tasks that share the same  dynamics, it is still challenging to transfer among tasks across domains.

\textbf{Vision Transformer.} With the great successes of Transformers ~\cite{vaswani2017attention} in NLP~\cite{devlin2018bert, radford2015unsupervised}, people apply them to solve computer vision problems. ViT~\cite{dosovitskiy2020image} is the first pure Transformer model introduced into the vision community and surpasses CNNs with large scale pretraining on the private JFT dataset~\cite{riquelme2021scaling}. DeiT~\cite{touvron2021training} trains ViT from scratch on ImageNet-1K~\cite{deng2009imagenet} and achieves better performance than CNN counterparts. Pyramid ViT (PVT)~\cite{wang2021pyramid} is the first hierarchical design for ViT, and proposes a progressive shrinking pyramid and spatial-reduction attention.
Swin Transformer~\cite{liu2021swin} computes attention within a local window and adopts shifted windows for communication aggregation. More recently, efficient transfer learning is also explored in for vision Transformer~\cite{bahng-2022-vp, jia-2022-vpt, chen2022adaptformer}. In this paper, we take the original ViT~\cite{dosovitskiy2020image} as the visual backbone with simple pooling layers, which are used to reduce the calculation burden, and more advanced structures may bring further gain.

%% file: src/2-settings.tex
\section{Preliminaries}

\textbf{Overview of Vision Transformer.}
The original Transformer~\cite{vaswani2017attention} tasks as input a 1D sequence of token embeddings. To handle 2D images, ViT~\cite{dosovitskiy2020image} splits an input image $\mathbf{x} \in \mathbb{R}^{H\times W\times C}$ to a sequence of flattened 2D patches $\mathbf{x}_p \in \mathbb{R}^{N\times(P^2 \cdot C)}$, where we let $H$ and $W$ denote the height and width of the image, $C$ the number of channels, $(P, P)$ the resolution of each image patch, and $N = \frac{HW}{P^2}$ is the number of flattened patches. After obtaining $\mathbf{x}_p$, ViT map it to $D$ dimensions with a trainable linear projection and uses this constant latent vector size $D$ through all of its layers. The output of this projection is named the patch embeddings.
\newcommand{\mbf}[1]{\mathbf{#1}}
\begin{equation}
        \mathbf{z}_0 = [ \mbf{x}_\text{class}; \, \mbf{x}^1_p \mbf{E}; \, \mbf{x}^2_p \mbf{E}; \cdots; \, \mbf{x}^{N}_p \mbf{E} ] 
\end{equation}
where $\mbf{E} \in \mathbb{R}^{(P^2 \cdot C) \times D}$ is the projection matrix. $\mbf{x}_\text{class}$ denotes the class token. 
The Transformer block~\cite{vaswani2017attention} includes alternating layers of multiheaded self-attention~(MHSA) and MLP blocks. Besides, Layernorm~(LN) is applied before every block, and residual connections after every block. The MLP utilizes two layers with a GELU~\cite{hendrycks2016gaussian} non-linearity. This process can be formulated as:
\newcommand{\op}[1]{\text{#1}}
\begin{equation}
\footnotesize
\begin{split}
    \mbf{z^\prime}_\ell &= \operatorname{MHSA}(\op{LN}(\mbf{z}_{\ell-1})) + \mbf{z}_{\ell-1}, \text{\quad} \ell=1\ldots L \\
    \mbf{z}_\ell &= \operatorname{MLP}(\op{LN}(\mbf{z^\prime}_{\ell})) + \mbf{z^\prime}_{\ell}, \text{\quad} \ell=1\ldots L\\
    \mbf{y} &= \operatorname{LN}(\mbf{z}_L^0))
\end{split}
\end{equation}
where $\mbf{y}$ denotes the image representation, which is encoded by the output of the class token at the last block.

To build up a Transformer block, an MLP~\cite{popescu2009multilayer} block with two linear transformations and GELU~\cite{hendrycks2016gaussian} activation are usually adopted to provide nonlinearity. Note that the dimensions of the parameter matrix will not change when the number of tokens increases, which is a key advantage of the Transformer handling variable-length inputs.

\textbf{Reinforcement Learning for Visual Control.} Reinforcement learning for visual control aims to learn the optimal policy given the observed images, and could be formulated as an infinite-horizon partially observable Markov decision process (POMDP) \cite{bellman1957mdp,kaelbling1998planning}. POMDP can be denoted by $\mathcal{M}=\langle\mathcal{O}, \mathcal{A}, \mathcal{P}, p_0, r, \gamma\rangle$, where $\mathcal{O}$ is the high-dimensional observation space (\ie, image pixels), $\mathcal{A}$ is the action space,  $\mathcal{P} = Pr(o_{t+1}|o_{\leq t},a_{t})$  represents the probability distribution over the next observation $o_{t+1}$ given the history of previous observations $o_{\leq t}$ and the current action $a_{t}$ and $p_0$ is the distribution of initial state. $r: \mathcal{O} \times \mathcal{A} \rightarrow \mathbb{R}$ is the reward function that maps the current observation and action to a scalar representing the reward, $r_t  = r(o_{\leq t}, a_t)$. The overall objective is to find the optimal policy $\pi^*$ to maximize the cumulative discounted return $E_{\pi}[\sum_{t=0}^{\infty} \gamma^t r_t | a_t \sim \pi(\cdot|s_{\leq t}), s_{t+1} \sim p(\cdot|s_{\leq t}, a_t), s_0 \sim p_0(\cdot)]$, where $\gamma \in [0, 1)$ is the discount factor, which is applied to pay more attention on recent rewards rather than future ones  and is  usually set to 0.99 in practice.

By stacking several consecutive image observations into a state, $s_t = \{o_t, o_{t-1}, o_{t-2}, \ldots\}$, the POMDP could be converted into an Markov Decision Process (MDP)~\cite{bellman1957mdp}, where information at the next time-step is determined by the information at current step, unrelated to those in the history. Thus, for a MDP process,  the transition dynamics can be refined as $p = Pr(s'_{t}|s_t,a_{t})$ representing the distribution of next state $s'_{t}$ given the current state $s_{t}$ and action $a_{t}$, and the reward function is refined as $r_t = r(s_t, a_t)$ similarly. 
In practice, three consecutive images are stacked into a state $s_{t}$ and as an MDP, the objective turns into finding the optimal policy, $\pi^{*}(a_t|s_t)$ to maximize the expected return.

%% file: src/3-method.tex
\section{Method}

In this section, we introduce our CtrlFormer for visual control in details. As shown in Figure \ref{fig:Overview}, the observation is first split into $N$ patches and mapped to $N$ tokens $[\mathbf{x}_p^1;\cdots;\mathbf{x}_p^N]$. Then CtrlFormer takes as inputs the image patches with an contrastive token $\mathbf{x}_\text{con}$ to improve the sample efficiency and $K$ policy tokens $[\mathbf{x}_{\pi}^1;\cdots;\mathbf{x}_{\pi}^K]$ and interactively encodes them with self-attention mechanism, leading to  representations $[\mathbf{z}_\text{con};\mathbf{z}_{\pi}^1;\cdots;\mathbf{z}_{\pi}^K;\mathbf{z}_{p}^1;\cdots;\mathbf{z}_{p}^N]$. Each task is assigned a policy token $x^{i}_{\pi}$, which is a randomly-initialized but learnable variable similar to the class token in conventional vision transformers.
In training, the policy token learns to abstract the characteristic of the corresponding task and the correlations across tasks via gradient back-propagation.
In inference, it serves as the query to progressively gather useful information from visual inputs and previously learned tasks through self-attention layers.
 To train the \model encoder, the representation of each policy token, $\mathbf{z}_{\pi}^{i}$ is utilized as the input of the task-specific policy network and Q-network in the downstream reinforcement learning algorithm, and the goal is to maximize the expected return for each task. The data-regularized method is used to reduce the variance of Q-learning and improve the robustness of the change of representation. Besides, a contrastive objective is applied on $\mathbf{z}_{\text{con}}$ to aid the training process, which significantly improves sample efficiency. The total loss is the sum of the reinforcement learning part and the contrastive learning part in equal proportion.
 
 In Section \ref{subsec: architecture}, we first introduce the details in CtrlFormer, and then a discussion is presented on how to transfer the learned representations to a new task in Section \ref{subsec: transfer}. The two objectives for training CtrlFormer, i.e., the policy learning problem and contrastive learning, are discussed in detail respectively in Section \ref{subsec: RL} and \ref{subsec: contrastive}.
 
 \begin{figure}[t]
\begin{center}
    \includegraphics[width=0.92\textwidth]{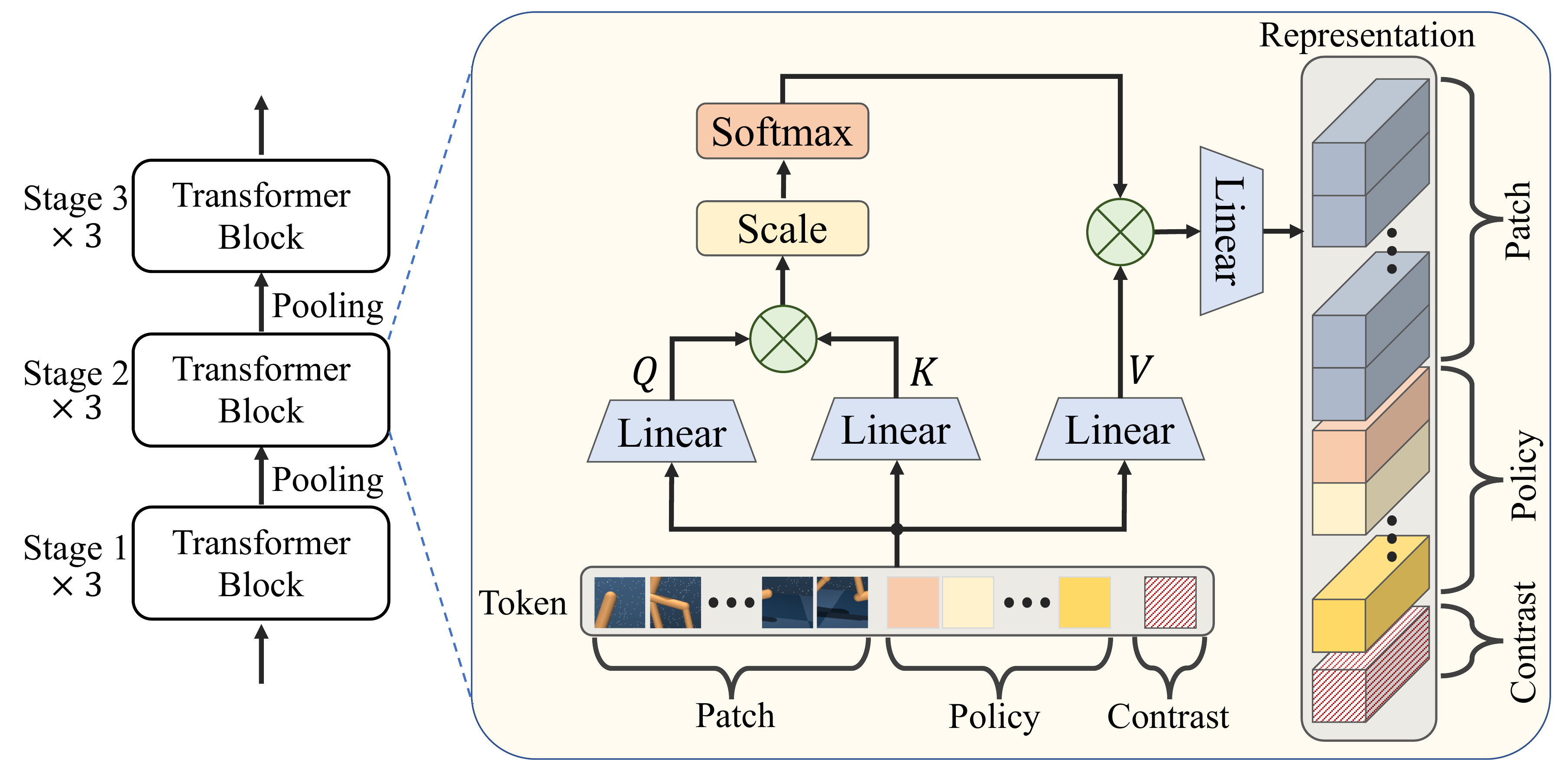}
\end{center}
\vspace{-10pt}
\caption{\textbf{The structure of \model.} 
    \model has a pyramid structure consisting of 3 stages and each stage has 3 blocks. The input image is split into several patches and these patches are mapped to a sequence of \textit{patch} tokens. The number of patch tokens is reduced to half by pooling operation between two stages. Each task has an independent \textit{policy} token. The number of policy tokens maintains through 3 stages. All tasks shares an additional \textit{constractive} token. The representations are learned by the self-attention mechanism. The output of the policy token is used for downstream reinforcement learning and the output of the contrastive token is used for contrastive learning.
    }\label{fig:overall-pipeline}
\vspace{-3pt}
\end{figure}

\subsection{The Architecture of \model }
\label{subsec: architecture}
In \model, as shown in Figure~\ref{fig:overall-pipeline},  each task has independent \verb|[policy]| token $\mathbf{x}_{\pi}^{i}$, which is similar to \verb|[class]| token in ViT~\cite{dosovitskiy2020image}.  Three consecutive frames of images are stacked into a 9-channel input tensor $\mathbf{x}_{p} \in \mathbb{R}^{H \times W \times 9}$.
We split the input tensor into $N=9HW/P^{2}$ patches with patch size $P \times P$ and then map it to a sequence of vectors  $\left\{\mathbf{x}_{p}^{1},\mathbf{x}_{p}^{2},\ldots,\mathbf{x}_{p}^{N}\right\}$.
Contrastive learning is learned together with reinforcement learning as an auxiliary task to improve the sample efficiency, which is assigned to an \verb|[contrastive]| token $\mathbf{x}_{\text{con}}$. Position embeddings $\mathbf{E}_{\text {pos}} \in \mathbb{R}^{N \times D}$, which is the same as those used in ~\cite{dosovitskiy2020image}, are added to the patch embeddings $\left\{\mathbf{x}_{p}^{i}\right\}_{i=1}^{N}$ to retain positional information.

Thus, the input of the transformer is
\begin{equation}
\vspace{-10pt}
\mathbf{z}_{\ell_0}=\left[\mathbf{x}_{\text{con}};\mathbf{x}_{\pi}^{1};\ldots;\mathbf{x}_{\pi}^{K}; \mathbf{x}_{p}^{1};\cdots; \mathbf{x}_{p}^{N} \right]+\mathbf{E}_{\text{pos}} 
\end{equation}

The blocks with multi-head self-attention~(MHSA) ~\cite{vaswani2017attention} and layer normalization~(LN) ~\cite{ba2016layer} could be formulated as 
\begin{equation}
\footnotesize
\begin{split}
\mathbf{z}_{\ell_{j}}^{\prime}
&=\operatorname{MHSA}\left(\operatorname{LN}\left(\mathbf{z}_{\ell_{j-1}}\right)\right)+\mathbf{z}_{\ell_{j-1}} \\
\mathbf{z}_{\ell_{j}} &=\operatorname{MLP}\left(\operatorname{LN}\left(z_{\ell_{j}}^{\prime}\right)\right)+\mathbf{z}_{\ell_{j}}^{\prime} \\
\end{split}
\end{equation}
where $\mathbf{z}_{l_{0}}$ is the input of transformer and $\mathbf{z}_{l_{j}}$
is the output of the $j$-th block.
To reduce the number of parameters needed to learn, we utilize a pyramidal structure. There are three stages in the pyramidal vision transformer, and the number of tokens decreases with the stages by a pooling layer. In the pooling layer, the token sequence with a length of $N$ is reshaped into $(H/P) \times (W/P)$ and is pooled by a 2$\times$2 filter with stride $(2,1)$ in the first stage and with stride $(1,2)$ in the second stage. After pooling, the tensor is flattened back to the token sequence with a length of $N$ to serve as the input of the next stage. More detailed structure of \model is introduced in Figure \ref{fig:detailed-structure} in Appendix \ref{app: detailed-structure}.

\subsection{Representation Transferring in New Task}
\label{subsec: transfer}
After learning the $K$-th previous task with policy token $\mathbf{z}_{\pi}^{K}$, we pad a new policy token $\mathbf{z}_{\pi}^{K+1}$, and the new task is learned with policy token $\mathbf{z}_{\pi}^{K+1}$, policy network $\pi^{K+1}(\cdot)$ and Q-networks $Q^{K+1}_{1}(\cdot,\cdot)$ and $Q^{K+1}_{2}(\cdot,\cdot)$. \model inherits the merit of the transformer model to resolve variable-length input sequences. The dimension of the \model's model parameters remains unchanged when the number of policy tokens changes, \textit{i.e.}, the weight dimension in self-attention is only related to the dimension of the token rather than the number of tokens. 
The parameter metrics  $\mathbf{W}^{q} \in \mathbb{R}^{D \times D}$,$\mathbf{W}^{k} \in \mathbb{R}^{D \times D}$ and $\mathbf{W}^{v} \in \mathbb{R}^{D \times D}$ are all remain its original dimension. When transferring to the $K$+1 task, the number of input tokens is increased from $N$+$K$+1 to $N$+$K$+2, hence $N$+$K$+2 output representations. Thus there is always a one-to-one correspondence between policy tokens and output representations. Both old and new tasks can be tackled within \model. 
The policy query $\mathbf{q}_{\pi} \in \mathbb{R}^{(K+1) \times D}$ and the key $\mathbf{k} \in \mathbb{R}^{(N+K+2) \times D}$ are calculated by
\begin{equation}
\begin{aligned}
&\mathbf{q}_{\pi}=\mathbf{z}_{\pi} \mathbf{W}^{q}, \mathbf{k}=\mathbf{z} \mathbf{W}^{k} \\
&\mathbf{z}=\left[\mathbf{z}_{con} ; \mathbf{z}_{\pi}^{1} ; \ldots ; \mathbf{z}_{\pi}^{K+1} ; \mathbf{z}_{p}^{1};  \ldots ; \mathbf{z}_{p}^{N}\right]\\
&\mathbf{z}_{\pi}=\left[\mathbf{z}_{\pi}^{1} ; \ldots ; \mathbf{z}_{\pi}^{K+1}\right]
\end{aligned}
\end{equation}

\subsection{Downstream Visual Reinforcement Learning Task}
\begin{figure}
\begin{center}
    \begin{subfigure}{0.49\textwidth}
      \includegraphics[width=\textwidth]{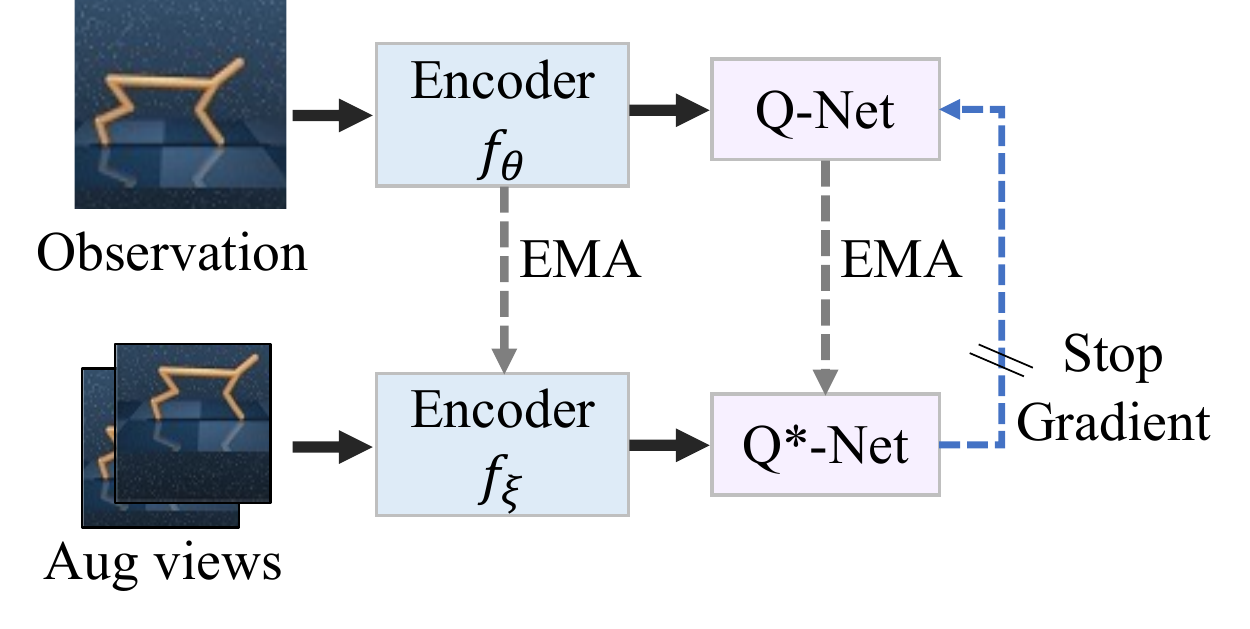}
      \caption{\footnotesize{Reinforcement learning.}}\label{RL_Branch}
    \end{subfigure}
    \begin{subfigure}{0.49\textwidth}
      \includegraphics[width=0.98\textwidth]{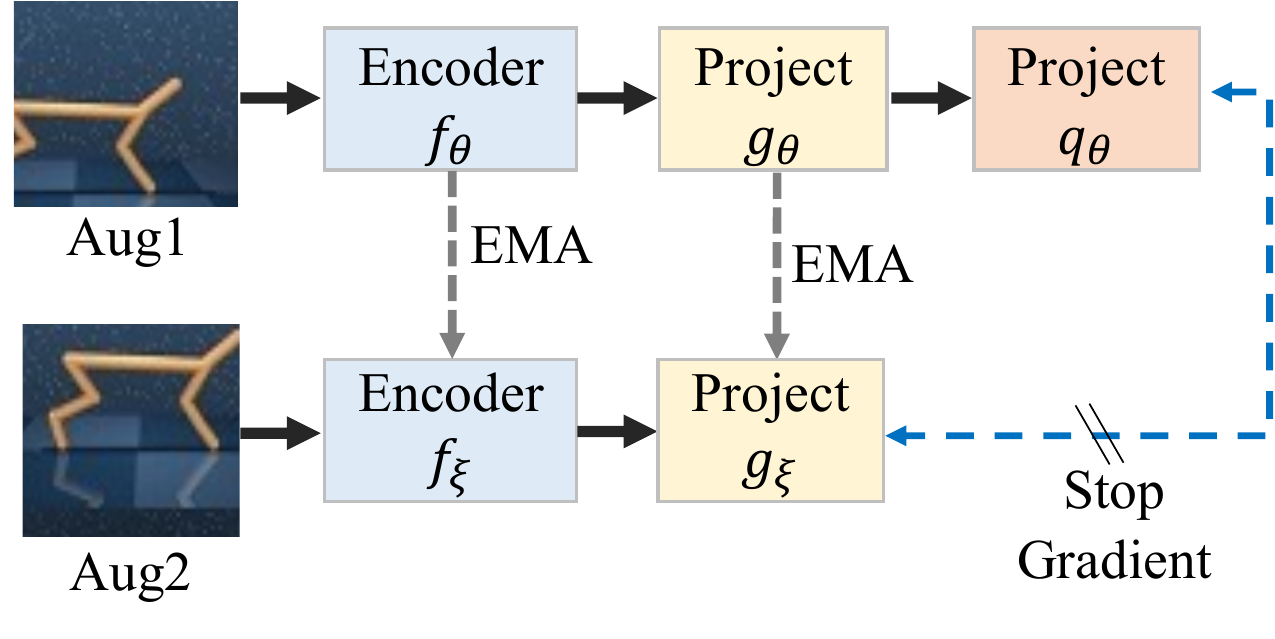}
      \caption{Contrastive learning.}\label{contrastive_branch }
    \end{subfigure}
\end{center}
\vspace{-10pt}
\footnotesize
\caption{\textbf{Downstream reinforcement learning with contrastive learning co-training.} Both reinforcement learning and contrastive learning are based on a set of momentum-updated learning frameworks, including online encoder $f_{\theta}$, Q-networks $Q(\cdot, \cdot)$ and online projector $g_{\theta}$, and target encoder $f_{\xi}$, target Q-network $Q^{*}(\cdot,\cdot)$ and target projector $g_{\xi}$, which are updated by EMA method. 
In the RL part, the observation is firstly encoded by the online encoder as the input of $Q(\cdot, \cdot)$. Different augmented views of the observation are encoded by the target encoder $f_{\xi}$ as the input of the $Q^{*}(\cdot, \cdot)$. The target Q-value is the mean of the value calculated by different views. The encoder and Q-networks are updated by the Critic loss calculated by the estimated Q-value and target Q-value. The objective of contrastive learning is to predict the projected representation from one augmented view of observation with the projected representation from another augmented view. }\label{fig:Downstream}
\end{figure}
\label{subsec: RL}
With the output $\mathbf{z}=[\mathbf{z}_{\text{con}};\mathbf{z}_{\pi}^{1};\ldots;\mathbf{z}_{\pi}^{K};\mathbf{z}_{p}^{1};\ldots;\mathbf{z}_{p}^{N}]$ of the \model,  we take $\mathbf{z}_{\pi}^{i}$ as the representation for the $i$-th task, which contains the task-relevant information gathered from all image tokens and policy tokens through the self-attention mechanism.
We utilize SAC ~\cite{haarnoja2018soft} as the downstream reinforcement learning algorithm to solve the optimal policy with the representation $\mathbf{z}_{\pi}^{i}$ for the $i$-th task, which is an off-policy RL algorithm that optimizes a stochastic policy for maximizing the expected trajectory returns. SAC learns a stochastic policy $\pi_\psi$ and critics $Q_{\phi_1}$ and $Q_{\phi_2}$ to optimize a $\gamma$-discounted maximum-entropy objective ~\cite{ziebart2008maxent}. We use the same structure of the downstream Q-network and policy network as DrQ ~\cite{kostrikov2020image}. More details about SAC are introduced in Appendix \ref{section:extended_background} and the detailed  structures of Q-network and policy network are introduced in Appendix ~\ref{app: detailed-structure}.  The parameters $\phi_j$ are learned by minimizing the Bellman error:
\begin{equation}\footnotesize\label{eq:qmsbe}
   \mathcal L (\phi_{j},\mathcal{B}) = \mathbb{E}_{t \sim \mathcal B} \left [\left ( Q_{\phi_j}(\mathbf{z}_{\pi}^{i},a) - \left( r+\gamma(1-d)\mathcal T  \right )\right )^2 \right]
\end{equation}
where $(\mathbf{z}_{\pi}^{i},a,{\mathbf{z}^{\prime}}_{\pi}^{i},r,d)$ is a tuple with current latent state $z_{\pi}^{i}$, action $a$, next latent state ${\mathbf{z}^{\prime}}_{\pi}^{i}$, reward $r$ and done signal $d$, $\mathcal{B}$ is the replay buffer, and $\mathcal T$ is the target, defined as:
\begin{equation}\footnotesize\label{eq:target}
   \mathcal{T}  = \left (\min_{i=1,2} Q^{*}_{\phi_j} ({\mathbf{z}^{\prime}}_{\pi}^{i},a') - \alpha \log \pi_\psi(a'|{\mathbf{z}^{\prime}}_{\pi}^{i})\right )
\end{equation}
In the target  \eqref{eq:target}, $Q^{*}_{\phi_j}$ denotes the exponential moving average (EMA)~\cite{holt2004forecasting} of the parameters of $Q_{\phi_j}$. Using the EMA has empirically shown to improve training stability in off-policy RL algorithms \citep{mnih2015human}. The parameter $\alpha$ is a positive entropy coefficient that determines the priority of the entropy maximization during the policy optimization.

As shown in Figure \ref{RL_Branch},
to improve the robustness of the policy network and Q-network and further reduce the variance of Q-learning,
we regularize the Q-target function $Q_{\phi_{j}}^{*}(\cdot,\cdot)$ by data augmentation as shown in  \eqref{eq:DRQ}
\vspace{-10pt}
\begin{equation}
\footnotesize
Q_{\phi_{j}}^{*}\left(\mathbf{z}_{\pi}^{i}, a\right) = \frac{1}{K} \sum_{m=1}^{K} Q_{\phi_{j}}^{*}\left(\mathbf{z}_{\pi}^{i,m}, a_{m}^{\prime}\right)
\vspace{-5pt}
\label{eq:DRQ}
\end{equation}
where $\mathbf{z}_{\pi}^{i,m}$ is encoded by a random augmented image input, $a_{m} \sim \pi\left(\cdot \mid \mathbf{z}_{\pi}^{i,m}\right)$.
Such a Q-regularization method allows the generation of several surrogate observations with the same $Q$-values, thus providing a mechanism to reduce the variance of $Q$-function estimation. 

While the critic is given by $Q_{\phi_{j}}$, the actor samples actions from policy $\pi_\psi$ and is trained by maximizing the expected return of its actions as in:
\begin{equation}\label{eq:actorloss}
   \mathcal L (\psi) = \mathbb{E}_{a \sim \pi} \left [ Q^\pi (\mathbf{z}_{\pi}^{i},a) - \alpha \log \pi_\psi (a|\mathbf{z}_{\pi}^{i}) \right ]
\end{equation}

\subsection{Contrastive Learning for Efficient Downstream RL}
\label{subsec: contrastive}
\label{sec:Contrastive}
Transformers are empirically proven hard to perform well when trained on insufficient amounts of data. However, reinforcement learning aims to learn behaviour with as little as possible interaction data, which is considered expensive to collect.
DeiT~\cite{touvron2021training} introduces a teacher-student strategy for data-efficient transformer training, which relies on an auxiliary distillation token, ensuring the student learns from the teacher through attention. By referencing this idea, we propose a sample-efficient co-training method with contrastive learning for transformer in reinforcement learning. 
The goal of the contrastive task is to learn a representation $\mathbf{z}_{\text{con}}$, which can then be used for downstream tasks. As shown in Figure \ref{contrastive_branch }, we use a momentum learning framework, which contains an online network and a target network that is updated by the exponential moving average~(EMA) method. The online network is defined by a set of weights $\netparams$ and is comprised of three stages: an encoder $f_\netparams$, a projector~$g_\netparams$, a predictor~$q_\netparams$. 
The target network, which is defined by a set of weights $\targetparams$, provides the regression targets to train the online network. As shown in Figure~\ref{contrastive_branch }, the observation $o$ is randomly augmented by two different image augmentation $t$ and $t^{\prime}$ respectively from two distributions of \imageaugmentation s $\mathcal{T}$ and $\mathcal{T}'$.
 Two \augview s $v \triangleq t(o)$ and $v' \triangleq t'(o)$ are produced from $o$ by applying respectively \imageaugmentation s $t\sim \mathcal{T}$ and $t'\sim \mathcal{T}'$. The details of the image augmentation implementation is introduced in Appendix\ref{augmentaion}.

From the first \augview~$v$, the online network outputs a representation $\mathbf{z}_{\text{con}} \triangleq f_\netparams(v)$ by the vision transformer and a projection $y_\netparams \triangleq g_\netparams(\mathbf{z}_{\text{con}})$. The target network outputs $z^{\prime}_{\text{con}_\targetparams} \triangleq f_\targetparams(v')$  and the target projection $y^{\prime}_\targetparams \triangleq g_\targetparams (z^{\prime}_{\text{con}_\targetparams})$ from the second \augview~$v'$, which are all stop-gradient. 
We then output a prediction $q_\netparams(y_\netparams)$ of $y'_\targetparams$ and $\ell_2$-normalize both $q_\netparams(y_\netparams)$ and $y'_\targetparams$ to $\overline{q_{\theta}}\left(y_{\theta}\right) \triangleq q_{\theta}\left(y_{\theta}\right) /\left\|q_{\theta}\left(y_{\theta}\right)\right\|_{2}$ and $\bar{y}_{\xi}^{\prime} \triangleq y_{\xi}^{\prime} /\left\|y_{\xi}^{\prime}\right\|_{2}$. 
The contrastive objective is defined by the mean squared error between the normalized predictions and target projections,
\begin{equation}
\footnotesize
\mathcal{L}_{\theta, \xi} \triangleq\left\|\overline{q_{\theta}}\left(y_{\theta}\right)-\bar{y}_{\xi}^{\prime}\right\|_{2}^{2}=2-2 \cdot \frac{\left\langle q_{\theta}\left(y_{\theta}\right), y_{\xi}^{\prime}\right\rangle}{\left\|q_{\theta}\left(y_{\theta}\right)\right\|_{2} \cdot\left\|y_{\xi}^{\prime}\right\|_{2}}.
\end{equation}
The updates of online network and target network are summarized as
\begin{equation}
\begin{aligned}
&\theta \leftarrow \operatorname{optimizer}\left(\theta, \nabla_{\theta} \mathcal{L}_{\theta, \xi}, \eta\right), \\
&\xi \leftarrow \tau \xi+(1-\tau) \theta,
\end{aligned}
\end{equation}
where $\mathrm{optimizer}$ is an optimizer and $\eta$ is a learning rate.

%% file: src/4-exp.tex
\section{Experiments}
In this section, we evaluate our proposed \model on multiple domains in DMControl benchmark~\cite{tassa2018deepmind}. We test the transferability among the tasks in the same domain and the tasks across different domains. Throughout these experiments, the encoder, actor, and critic neural networks are trained using the Adamw optimizer~\cite{loshchilov2017decoupled} with the learning rate $lr=10^{-4}$ and a mini-batch size of 512. The soft target update rate $\tau$ of the critic is 0.01, and target network updates are made every 2 critic updates (same as in DrQ~\cite{kostrikov2020image}). The full set of parameters is in Appendix~\ref{app:Detailed_Training}. The Pytorch-like pseudo-code is provided in Appendix~\ref{app:Pseudo-code}.
\begin{table}[h]
\vspace{-1mm}
\begin{center}
    \footnotesize
    \setlength{\tabcolsep}{8pt}
     \renewcommand{\arraystretch}{1.0}
    \begin{tabular}{l|c c}
    \Xhline{1pt}
     Domain                        & Task 1 & Task 2 \\
     \Xhline{1pt}
     \textbf{\walkercolor{Walker}} & \texttt{stand} & \texttt{walk} \\
     \textbf{\cartpolecolor{Cartpole}} & \texttt{swingup} & \texttt{swingup-sparse} \\
     \textbf{\reachercolor{Reacher}}   & \texttt{easy}  & \texttt{hard} \\
     \textbf{\fingercolor{Finger}}     & \texttt{turn-easy} & \texttt{hard} \\
     \Xhline{1pt}
    \end{tabular}
\end{center}
\vspace{-3mm}
\caption{Domains and the corresponding tasks.}\label{tab:task-domain}
\vspace{-3mm}
\end{table}
\vspace{-10pt}
\input{src/table1}

\input{src/table2}

\subsection{Benchmark}
The DeepMind control suite is a set of continuous control tasks and has been widely used as the benchmark for visual control~\cite{tassa2018deepmind}. We mainly test the performance of \model in 4 typical domains from DeepMind control suite. 
The dimensions of action space are the same for tasks within the same domain and different across domains.
In order to evaluate the performance of \model among the tasks in same-domain and cross-domain scenarios, we conduct extensive experiments on multiple domains and tasks, as shown in Table~\ref{tab:task-domain}.
The detailed introduction of the domains and tasks we use is in Appendix \ref{app:dmc_description}.
\subsection{Baselines}
We mainly compare \model with Dreamer~\cite{hafner2019dream}, DrQ~\cite{kostrikov2020image} and SAC~\cite{haarnoja2018sac} whose the representation encoded by a pre-trained ResNet~\cite{he2016deep}. Dreamer is a representative method with a latent variable model for state representation transferring, achieving the state-of-the-art performance for model-based reinforcement learning. To extract useful information from historical observations, it encodes the representation by a recurrent state space model~(RSSM)~\cite{hafner2019learning}. DrQ is the state-of-the-art model-free algorithm on DMControl tasks, which surpasses other model-free methods with task-specific representations~(\ie, methods update the encoder with the actor-critic gradients), such as CURL~\cite{laskin2020curl} and SAC-AutoEncoder~\cite{yarats2019improving}. 
To test the transferability of  DrQ, every task is assigned a specific head for the Q-network and the policy network, and all tasks share a CNN network to extract state representations. We utilize ResNet-50 ~\cite{he2016deep} as encoder and pre-trained it on the ImageNet~\cite{deng2009imagenet} to test the performance of task-agnostic representation for reinforcement learning.

\subsection{Transferability}

\noindent\textbf{Settings.} In all transferability testing experiments, the agent first learns a previous task, then adds a new policy token,  uses the previous policy token as the initialization of the current task policy token, and then starts learning a new task. Finally, we retested the score (average episode return) of the old task using the latest Encoder and the Actor and Critic networks corresponding to the old task after learning the new task. 

\textbf{Transferring in the same domain.} In the same domain, the dynamics of the agents from different tasks are similar, while the sizes of the target points and the characteristic of reward (sparse or dense) might be different.
We design an experimental pipeline to let the agent first learn an easier task and then transfer the obtained knowledge to a harder one in the same domain. The results are summarized in Table~\ref{table:Transferring_tasks_in_same_domain}. Compared to baselines, the sample efficiency of \model in the new task is improved significantly after transferring the state representation from the previous task. 
Taking Table~\ref{table:Transferring_tasks_in_same_domain-6} and Figure~\ref{fig:trans} as examples, \model achieves an average episode return of \texttt{$769_{\pm{34}}$} with representation transferring using 100$k$ samples, while learning from scratch is totally failed using the same number of samples.
Furthermore, when retesting on the previous task after transferring to new task, as shown in Figure~\ref{fig:maintain}, \model does not show an obvious decrease. However, the previous task's performance of DrQ is damaged significantly after learning the new task. 
Moreover, When learning from scratch, the sample efficiency of \model is also comparable to the DrQ with multi-heads. Although Dreamer shows promising transferability in the domain like \walkercolor{Walker}, where different tasks share the same dynamics, it transfers poorly in domains like \reachercolor{Reacher}, where the dynamics are different among tasks.
The representation encoded by a pre-trained ResNet-50 shows same performance among tasks but shows lower sample efficiency compared with \model and DrQ. 

\textbf{Transferring across domains.} Compared to the tasks in the same domain whose learning difficulty is readily defined in the DMControl benchmark, it is not easy to distinguish which task is easier to learn for tasks from different domains.
Thus, we test the bi-directional transferability in different domains, \ie, transferring from \fingercolor{Finger} domain to \reachercolor{Reacher} domain and transferring by an inverse order.
As shown in Table~\ref{table:cross_domain}, the \model has the best transferability and does not show a significant decrease after fine-turning on the new task. In contrast, the performance of DrQ decreases significantly after learning a new task and has worse performance damage than in the same domain because of a large gap across different domains.

\textbf{Sequential transferring among more tasks.} In order to test whether the improvement by state representation transfer is more significant with the increase of the number of tasks learned, we test the performance of \model in four tasks from easy to difficult sequentially and further compare the transfer leaning via \model to the method that learns the four tasks simultaneously. 
As shown in Table~\ref{tab:4tasks}, the method of sequentially learning four tasks via \model  has the best sample efficiency. Furthermore, we can find that transferring state representation from more tasks performs better than transferring from only one task. \model can surpass the performance of learning from scratch at 500$k$ while only using 100$k$ samples under this setting.

\begin{figure}[t]
\vspace{-10pt}
\begin{center}
\includegraphics[width=0.8\textwidth]{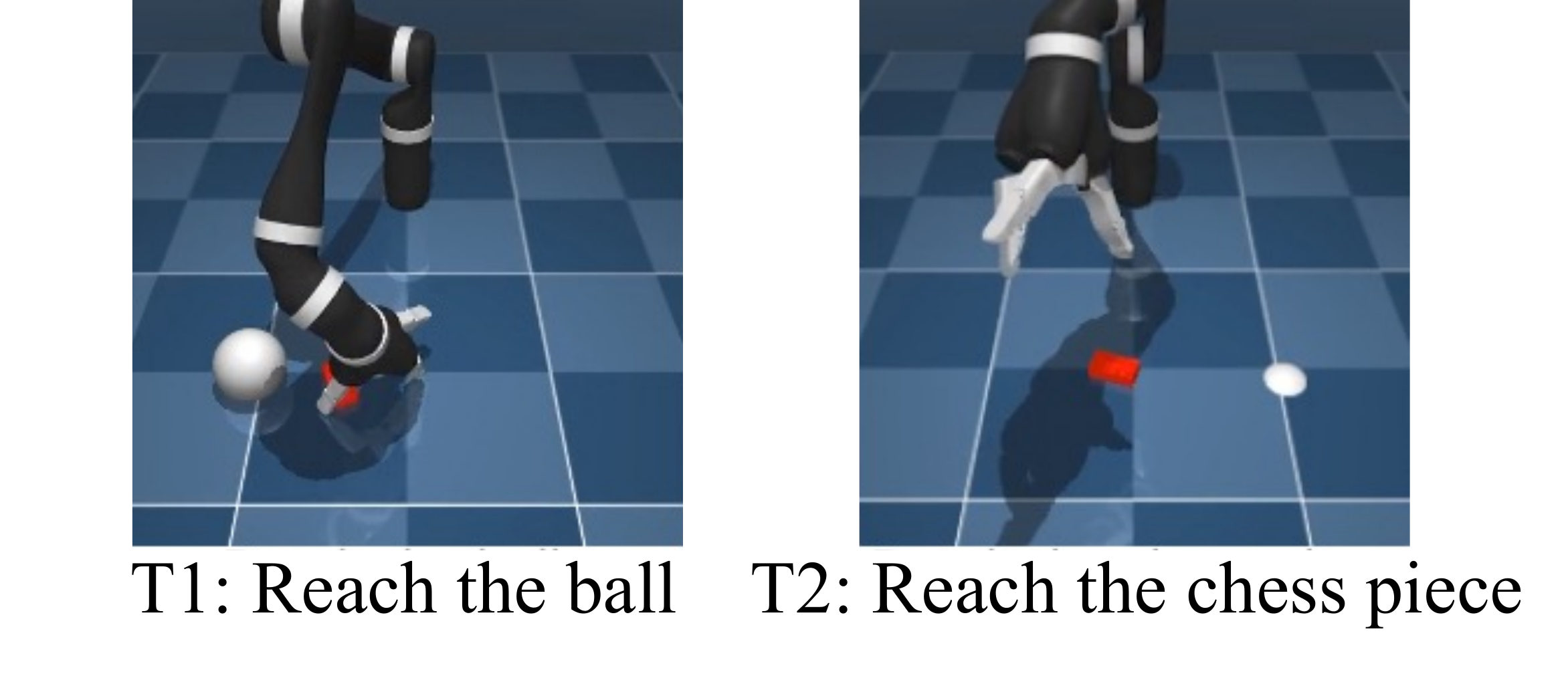}
\end{center}
\vspace{-15pt}
\caption{Tasks in  DeepMind manipulation}
\label{fig:mani-env}
\end{figure}

\begin{table}[ht]
\begin{center}
    \footnotesize
\setlength{\tabcolsep}{5pt}
\renewcommand{\arraystretch}{1.0}
    \begin{tabular}{l | c cc c}
\Xhline{1.0pt}
 Method & \multicolumn{4}{c}{Task 0 $\rightarrow$ Task 1  $\rightarrow$  Task 2 $\rightarrow$ Task 3} \\
 \Xhline{1.0pt}
  Scratch (100$k$) & \onepm{967}{27}& \onepm{869}{61} & \onepm{759}{48} & 0 \\
 Train together (100$k$) & \onepm{433}{23}&\onepm{143}{34} & \onepm{310}{41}  & $0$ \\
   \cellcolor{codegray}\model (100$k$) & \cellcolor{codegray}\onebfpm{967}{27}&\cellcolor{codegray}\onebfpm{981}{29} &\cellcolor{codegray}\onebfpm{988}{36} & \cellcolor{codegray}\onebfpm{853}{69} \\
   \hline
    Scratch (500$k$) & \onepm{995}{18}& \onepm{949}{44} & \onepm{846}{25} & \onepm{671}{81} \\
  Train together (500$k$) & \onepm{947}{32}&\onepm{942}{53} & \onepm{632}{44}  & \onepm{40}{15} \\
\cellcolor{codegray}\model (500$k$) & \cellcolor{codegray}\onebfpm{995}{18}&\cellcolor{codegray}\onebfpm{1000}{0} &\cellcolor{codegray}\onebfpm{992}{26} & \cellcolor{codegray}\onebfpm{878}{64} \\
\Xhline{1.0pt}
\end{tabular}
\end{center}
\vspace{-10pt}
    \caption{\textbf{Performance comparison with a series tasks.} Tasks 0-3 are \mono{balance}, \mono{balance\_sparse}, \mono{swingup} and  \mono{swingup\_sparse}, from \cartpolecolor{Cartpole} domain.}
    \label{tab:4tasks}
\end{table}
As shown in Figure \ref{fig:mani-env}, we also develop an experiment on the DeepMind manipulation benchmark \cite{tunyasuvunakool2020dm_control} to further show the advantage of CtrlFormer, which provides a Kinova robotic arm and a list of objects for building manipulation tasks. We test the performance of \model in 2 tasks: (1) Reach the ball: push the small red object near the white ball by the robot arm; (2) Reach the chess piece: push the small red object near the white chess piece by the robot arm. As shown in Table 2,  CtrlFormer surpasses DrQ in terms of both transferability and sample efficiency when learning from scratch. 
\begin{table}[t]
\footnotesize
    \begin{tabular}{l|c|c}
        \hline & DrQ & CtrlFormer \\
        \hline Scratch Task1(500k) & $154_{\pm 41}$ & $\textbf{175}_{\pm 63}$ \\
        \hline Retest Task1 & $87_{\pm 33}$ & $\textbf{162}_{\pm 75}$ \\
        \hline Scratch Task2(500k) & $141_{\pm 47}$ & $\textbf{164}_{\pm 33}$ \\
        \hline Transfer T1 to T2 (100k) & $73_{\pm 48}$ & $\textbf{116}_{\pm 34}$ \\
        \hline
    \end{tabular}
    \vspace{-13pt}
    \caption{Performance comparison on DeepMind manipulation.}
\end{table}

\begin{figure*}[ht]
    \centering
    \label{fig:Visualization}
        \begin{minipage}{0.31\linewidth}
    \begin{subfigure}{\linewidth}
        \centering
        \includegraphics[width=\linewidth]{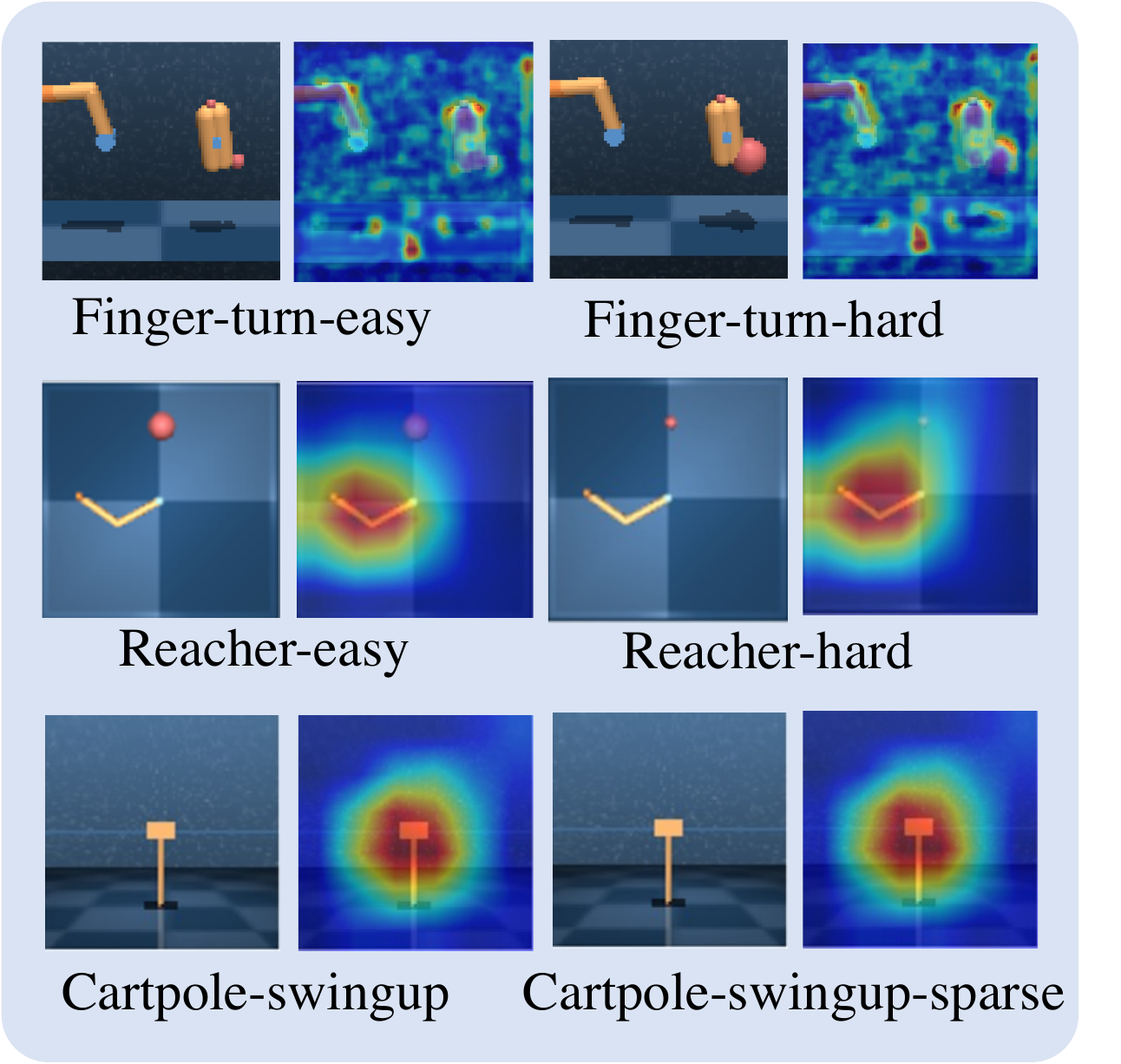}
        \caption{Visualization of Pre-trained ResNet}
        \label{fig:visual_res}
    \end{subfigure}
    \end{minipage}~~
        \begin{minipage}{0.31\linewidth}
    \begin{subfigure}{\linewidth}
        \centering
        \includegraphics[width=\linewidth]{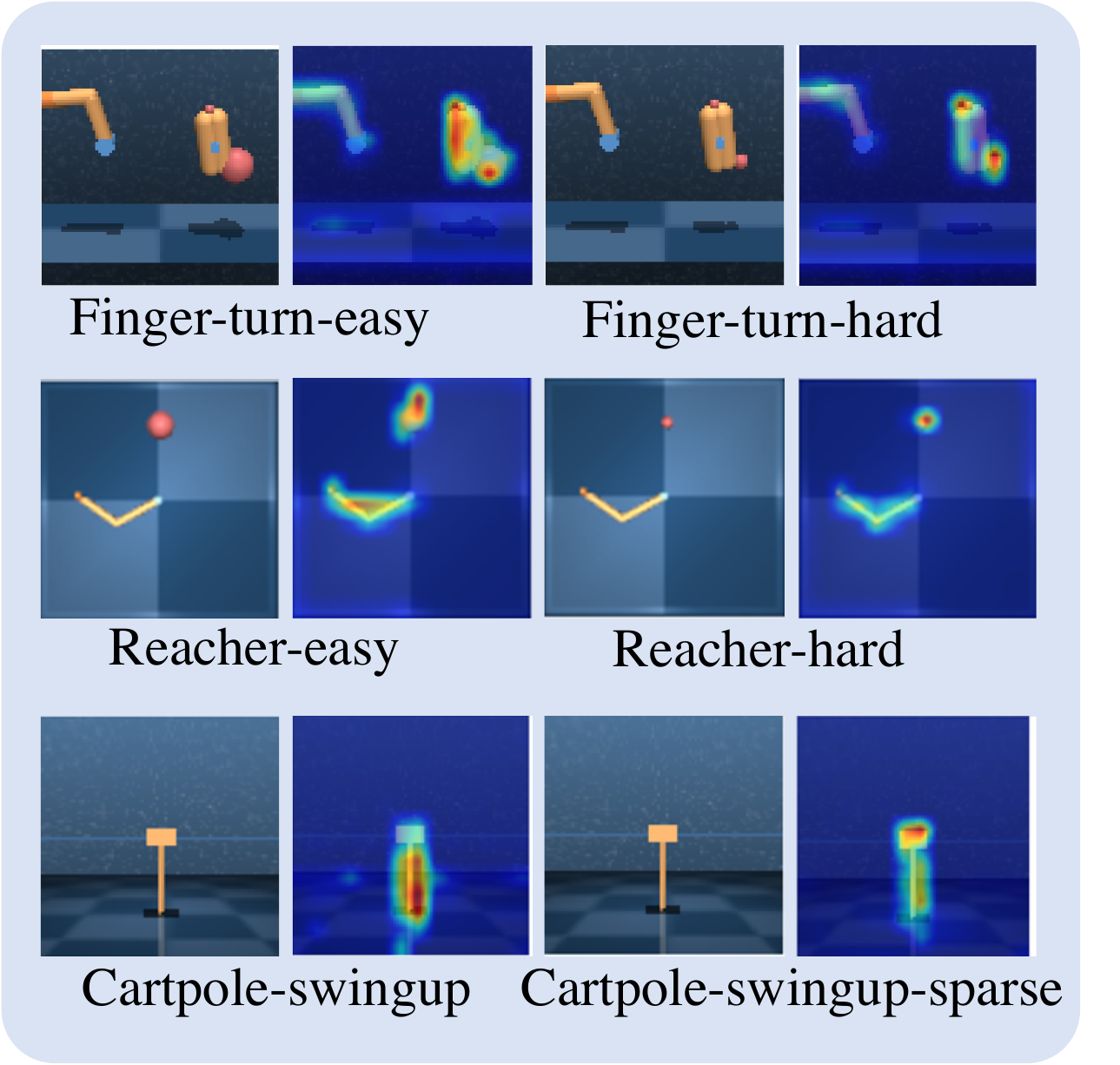}
        \caption{Visualization of \model}
        \label{fig:visual_control}
    \end{subfigure}
    \end{minipage}~~
    \begin{minipage}{0.295\linewidth}
    \begin{subfigure}{\linewidth}
        \centering
        \includegraphics[width=\linewidth]{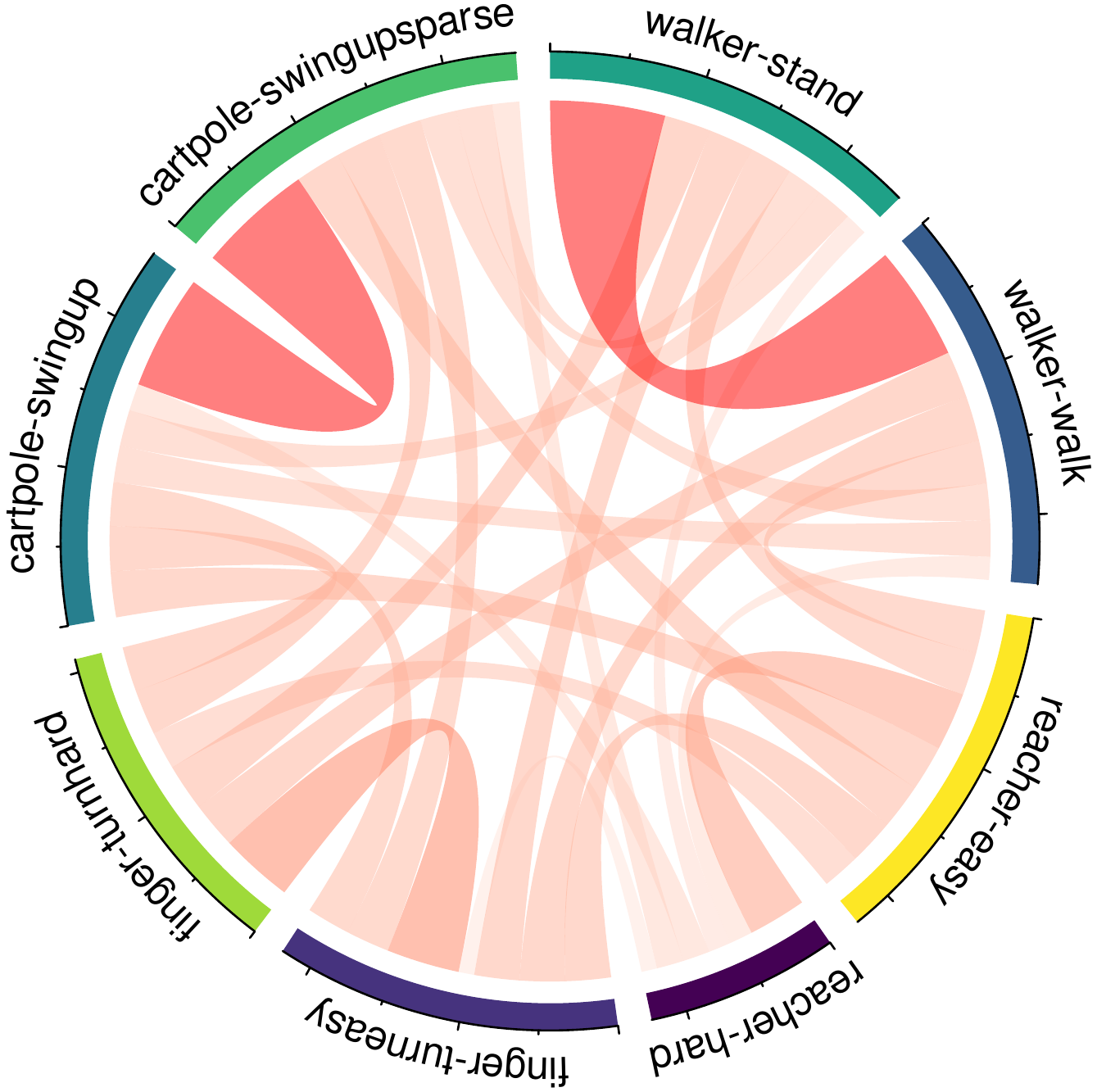}
        \caption{Visualized similarity between tasks}
        \label{fig:cos_similarity_dmc}
    \end{subfigure}%
    \end{minipage}
    \vspace{-15pt}
    \caption{Visualization of the attention on the input image and the similarity of different tasks in DMControl benchmark.}
    \label{fig:visualization}
\end{figure*}

\subsection{Visualization}
\textbf{Visualization of the similarity of different tasks.} We visualize the cosine similarity of policy tokens between tasks in Figure~\ref{fig:cos_similarity_dmc}, which reflects the similarity of the visual representation between different tasks.  The width and color of the bands in Figure \ref{fig:cos_similarity_dmc} represents the similarity of the representations between two tasks. The thicker the line and the darker the color, the higher the similarity between the two tasks.
As shown in Figure \ref{fig:cos_similarity_dmc},  the similarity between \walkercolor{Walker}~(\mono{stand}) and \walkercolor{Walker}~(\mono{walk}) and between \cartpolecolor{Cartpole}~(\mono{swingup}) and \cartpolecolor{Cartpole}~(\mono{swingupsparse}) in the same domain is the top two strongest, indicating that it has better feature transfer potential. This is consistent with our test results,  \model significantly improves the sample efficiency after transferring the state representation.
For the cross-domain, the similarity between different tasks is quite different, for example, the representation similarity between \fingercolor{finger}~(\mono{turn-easy}) and \reachercolor{reacher}~(\mono{easy}) is significantly higher than the similarity between finger~(turn-easy) and \reachercolor{reacher}~(\mono{hard}). This is because, in the \reachercolor{reacher}~(\mono{hard}) task, the size of the target point is quite small, and the model pays too much attention to the target point and relatively less attention to the rod, while the finger task focuses more on the rod control.

\begin{figure}
\begin{center}
    \begin{subfigure}{0.49\textwidth}
      \includegraphics[width=\textwidth]{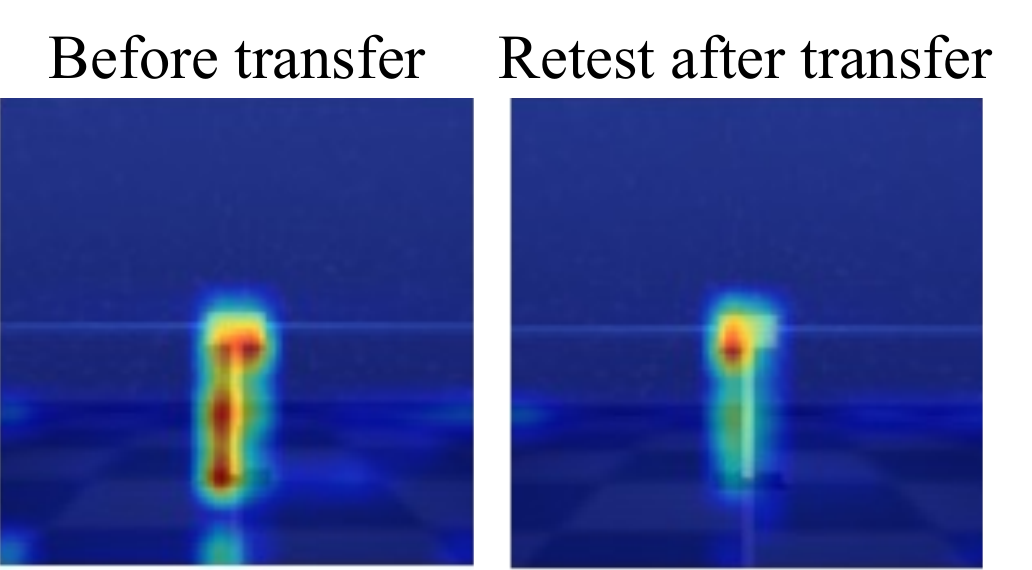}
      \caption{DrQ}\label{fig:cam-drq}
    \end{subfigure}
    \begin{subfigure}{0.49\textwidth}
      \includegraphics[width=\textwidth]{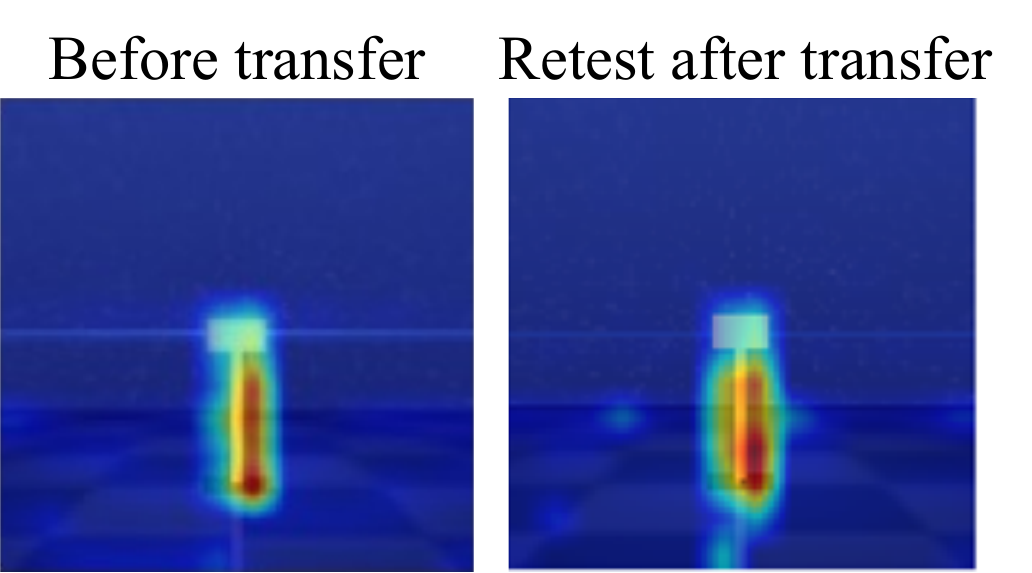}
      \caption{\model}\label{fig:cam-ctrlf}
    \end{subfigure}
\end{center}
\vspace{-18pt}
\caption{
\textbf{Comparison of the attention map change before and after the transferring.}  
Figure~\ref{fig:cam-drq} shows that DrQ fails to pay attention to some key areas after transferring. 
In contrast, our \model~(Figure~\ref{fig:cam-ctrlf}) has consistent attention areas before and after transferring, demonstrating its superior property of transferring without catastrophic forgetting.}
\vspace{-6pt}
\label{fig:BE_AF}
\end{figure}

\begin{table}[t]
    \centering
\setlength{\tabcolsep}{2pt}
\renewcommand{\arraystretch}{1.0}
    \begin{tabular}{l | c c|c c}
\Xhline{1.0pt}
\multirow{2}{*}{ \model } & \multicolumn{2}{c|}{w/o contrastive} & \multicolumn{2}{c}{w/ contrastive } \\
& $100 k$ & $500 k$ & $100 k$ & $500 k$ \\
\Xhline{1.0pt}
Cartpole(swingup) & \onepm{391}{42} & \onepm{835}{29}  & \onebfpm{759}{48} & \onebfpm{846}{25} \\
 Cartpole(swingup-sp) & 0 & \onepm{137}{54} & 0 & \onebfpm{671}{81}\\
 Reacher(easy) & \onepm{622}{38} & \onepm{917}{64}& \onebfpm{642}{42}& \onebfpm{973}{53}\\
 Reacher(hard) & \onepm{49}{15}&\onepm{234}{67}  &\onebfpm{104}{48}  &\onebfpm{548}{131}  \\
 Walker(stand) & \onepm{865}{33}   &\onepm{930}{44} & \onebfpm{877}{42} & \onebfpm{954}{38} \\
 Walker(walk) & \onepm{406}{73}& \onepm{708}{45} & \onebfpm{593}{52} &\onebfpm{903}{43} \\
 Finger(turn-easy) & \onepm{15}{7} & \onepm{344}{38}  & \onebfpm{281}{67}  & \onebfpm{493}{35} \\
 Finger(turn-hard) & \onepm{99}{44} & \onepm{136}{38} &\onebfpm{197}{78} & \onebfpm{344}{47}\\
\Xhline{1.0pt}
\end{tabular}
    \caption{Ablation study on the effectiveness of co-training with contrastive learning. }
    \label{tab:ablation_study}
\vspace{-12pt}
\end{table}

\textbf{Visualization of the attention on input image:} we visualize the attention of \model and pre-trained ResNet-50 on the input image by Grad-CAM ~\cite{selvaraju2016grad}, which is a typical visualization method, more details are introduced in Appendix \ref{app:Visual}. As shown in Figure \ref{fig:visual_res}, the attention of ResNet-50 is disturbed by a lot of things unrelated to the task, which is the key reason why it has low sample efficiency in both Table \ref{table:Transferring_tasks_in_same_domain} and Table \ref{table:cross_domain}. The attention of \model is highly correlated with the task, and different policy tokens learn different attention to the input image. The attention learned from similar tasks has similarities, but each has its own emphasis. Moreover, we also compare the attention map change on the old task before and after transferring in Figure~\ref{fig:BE_AF}, the attention map of \model is not obviously changed, while the attention map of CNN-based model used in DrQ changed obviously, which provides its poor retesting performance with a reasonable explanation.

\subsection{Ablation Study} 
To illustrate the effect of the co-training method with contrastive learning, we compare the sample efficiency of \model and \model without co-training, as shown in Table \ref{tab:ablation_study}, \model shows higher sample efficiency, proving the effectiveness of the proposed co-training for improving the sample efficiency of the transformer-based model.  

In conclusion, \model surpasses the baselines and shows great transferability for visual control tasks. 
The experiments in DMControl benchmark illustrated that \model has great potential to model the correlation and irrelevance between different tasks, which improves the sample efficiency significantly and avoids catastrophic forgetting.

%% file: src/table1.tex
\begin{table*}[ht]
\captionsetup{singlelinecheck = false, justification=justified}
\centering
\subfloat[\qquad  \textbf{Left:} Learn old task in \textbf{\cartpolecolor{Cartpole}}~(\mono{swingup})\qquad \qquad \qquad~\textbf{Right:} Transfer to new task \textbf{\cartpolecolor{Cartpole}}~(\texttt{swingup-sparse})
\label{table:Transferring_tasks_in_same_domain-6}
]{
\centering
\begin{minipage}{0.49\linewidth}{\begin{center}
\footnotesize
\setlength{\tabcolsep}{5pt}
\renewcommand{\arraystretch}{1.0}
\begin{tabular}{l | c c | c c}
\Xhline{1.0pt}
\multirow{2}{*}{Method}& \multicolumn{2}{c|}{Learn from scratch}& Retest after \\
&  100$k$ & 500$k$ & new task fine-tune \\
\Xhline{1.0pt}
DrQ        & \onepm{549}{36} & \onebfpm{854}{22} & \onepm{373}{24} \\
Dreamer    & \onepm{326}{27} & \onepm{762}{27}   & \onepm{704}{33} \\
Resnet+SAC & \onepm{192}{19} & \onepm{357}{85}   & \onepm{357}{85} \\
\cellcolor{codegray}CtrlFormer & \cellcolor{codegray} \onebfpm{759}{48} & \cellcolor{codegray}\onepm{846}{25} & \cellcolor{codegray}\onebfpm{842}{22} \\
\Xhline{1.0pt}
\end{tabular}
\end{center}}
\end{minipage}
\begin{minipage}{0.49\linewidth}{\begin{center}
\footnotesize
\setlength{\tabcolsep}{5pt}
\renewcommand{\arraystretch}{1.0}
\begin{tabular}{l | c c | c c}
\Xhline{1.0pt}
\multirow{2}{*}{Method}& \multicolumn{2}{c|}{Learn from scratch} & \multicolumn{2}{c}{Learn with transfer} \\
 & 100$k$& 500$k$ & 100$k$ & 500$k$ \\
\Xhline{1.0pt}
DrQ         &  0        & \onepm{505}{335} & 0  & \onepm{76}{41} \\
Dreamer     & \onepm{8}{4}     & \onepm{376}{214} & 0  & \onepm{589}{122}  \\
Resnet+SAC  &  0               & 0                & 0  & 0 \\
\cellcolor{codegray}CtrlFormer &  \cellcolor{codegray}0 & \cellcolor{codegray}\onebfpm{671}{81} &
\cellcolor{codegray}\onebfpm{769}{34}  & \cellcolor{codegray}\onebfpm{804}{26} \\ 
\Xhline{1.0pt}
\end{tabular}
\end{center}}\end{minipage}
}

\subfloat[\qquad \textbf{Left:}
Learn old task in \textbf{\fingercolor{Finger}}~(\mono{turn-easy}) \qquad \qquad \qquad \textbf{Right:}
Transfer to new task \textbf{\fingercolor{Finger}}~(\mono{turn-hard})
]{
\centering
\begin{minipage}{0.49\linewidth}{\begin{center}
\footnotesize
\setlength{\tabcolsep}{5pt}
\renewcommand{\arraystretch}{1.0}
\begin{tabular}{l | c c | c c}
\Xhline{1.0pt}
\multirow{2}{*}{Method}  & \multicolumn{2}{c|}{Learn from scratch} &Retest after \\
& 100$k$ & 500$k$ & new task fine-tune \\
\hline
 DrQ        &  \onebfpm{346}{33} & \onepm{448}{65}  & \onepm{300}{42} \\
 Dreamer    &  \onepm{25}{18}    & \onepm{245}{159} & \onepm{182}{34} \\
 Resnet+SAC &  \onepm{298}{17}   & \onepm{300}{29}  & \onepm{300}{29} \\
\cellcolor{codegray} \model & \cellcolor{codegray}\onepm{281}{67} & \cellcolor{codegray}\onebfpm{493}{35}  & \cellcolor{codegray}\onebfpm{475}{43} \\
\Xhline{1.0pt}
\end{tabular}
\end{center}}
\end{minipage}
\begin{minipage}{0.49\linewidth}{\begin{center}
\footnotesize
\setlength{\tabcolsep}{5pt}
\renewcommand{\arraystretch}{1.0}
\begin{tabular}{l | c c | c c}
\Xhline{1.0pt}
\multirow{2}{*}{Method}  & \multicolumn{2}{c}{Learn from scratch} & \multicolumn{2}{|c}{Learn with transfer} \\
 & 100$k$ & 500$k$ & \footnotesize{$100 k$} & \footnotesize{$500 k$} \\
\Xhline{1.0pt}
\footnotesize{DrQ} &  $8_{\pm 24}$ & $274 _{\pm 137}$ & $133_{\pm 26}$ & $455_{\pm 34}$ \\
 \footnotesize{Dreamer}  & $0$ & $17_{\pm 9}$ & $0$ & $38_{\pm 18}$ \\
\footnotesize{Resnet+SAC}  & $0$ & $17_{\pm 10}$ &$0$ & $17_{\pm 10}$ \\
\cellcolor{codegray}\footnotesize{CtrlFormer} &  \cellcolor{codegray}\footnotesize{$\textbf{197}_{\pm 78}$} & \cellcolor{codegray}\footnotesize{$\textbf{344}_{\pm 47}$} &
\cellcolor{codegray}\footnotesize{$\textbf{294}_{\pm 37}$} & \cellcolor{codegray}\footnotesize{$\textbf{569}_{\pm 32}$} \\ 
\Xhline{1.0pt}
\end{tabular}
\end{center}}\end{minipage}
}
\vspace{3pt}
\subfloat[\qquad \textbf{Left:}
Learning old task in \textbf{\reachercolor{Reacher}}~(\mono{easy}) \qquad \qquad \qquad \qquad \textbf{Right:} Transfer to new task \textbf{\reachercolor{Reacher}}~(\mono{hard})
]{
\centering
\begin{minipage}{0.49\linewidth}{\begin{center}
\footnotesize
\setlength{\tabcolsep}{5pt}
\renewcommand{\arraystretch}{1.0}
\begin{tabular}{l | c c | c c}
\Xhline{1.0pt}
\multirow{2}{*}{Method} & \multicolumn{2}{c|}{ \footnotesize {Learn from scratch} } &\footnotesize {Retest after} \\
& \footnotesize {$100 k$} & \footnotesize {$500 k$} & \footnotesize {new task fine-tune} \\
\Xhline{1.0pt}
 \footnotesize{DrQ} &  \footnotesize{$558_{ \pm 38}$} &  \footnotesize{$971_{ \pm 27}$} &  \footnotesize{$243_{ \pm 52}$} \\
 \footnotesize{Dreamer} &  \footnotesize{$314_{ \pm 155}$} &  \footnotesize{$793_{ \pm 164}$} &  \footnotesize{$485_{ \pm 67}$} \\
 \footnotesize{Resnet+SAC} &  \footnotesize{$322_{ \pm 285}$} &  \footnotesize{$382_{ \pm 299}$} &  \footnotesize{$382_{ \pm 299}$}\\
  \cellcolor{codegray}\footnotesize{CtrlFormer} &  \cellcolor{codegray}\footnotesize{$\textbf{642}_{\pm 42}$} &  \cellcolor{codegray}\footnotesize{$\textbf{973}_{ \pm 53}$} &  \cellcolor{codegray}\footnotesize{$\textbf{906}_{ \pm 31}$} \\
\Xhline{1.0pt}
\end{tabular}
\end{center}}\end{minipage}
\begin{minipage}{0.49\linewidth}{\begin{center}
\footnotesize
\setlength{\tabcolsep}{5pt}
\renewcommand{\arraystretch}{1.0}
\begin{tabular}{l | c c | c c}
\Xhline{1.0pt}
\multirow{2}{*}{Method} & \multicolumn{2}{c}{ \footnotesize{Learn from scratch} } & \multicolumn{2}{c}{ \footnotesize{Learn with transfer} } \\
 & \footnotesize{$100 k$} & \footnotesize{$500 k$} & \footnotesize{$100 k$} & \footnotesize{$500 k$} \\
\Xhline{1.0pt}
\footnotesize{DrQ} &  \footnotesize{$\textbf{194}_{\pm 84}$} &  \footnotesize{$\textbf{616}_{\pm 274}$} & \footnotesize{$96_{\pm 43}$} & \footnotesize{$524_{\pm 68}$} \\
 \footnotesize{Dreamer} &  \footnotesize{$13_{\pm 32}$} & \footnotesize{$115_{\pm 98}$} & \footnotesize{$63_{\pm 07}$} & \footnotesize{$148_{\pm 12}$} \\
\footnotesize{Resnet+SAC} & \footnotesize{$26_{\pm 4}$} & \footnotesize{$31_{\pm 12}$} & \footnotesize{$26_{\pm 4}$} & \footnotesize{$31_{\pm 12}$} \\
\cellcolor{codegray}\footnotesize{CtrlFormer} &  \cellcolor{codegray}\footnotesize{$104 \pm 48$} & \cellcolor{codegray}\footnotesize{$548_{\pm 131}$}&
\cellcolor{codegray}\footnotesize{$\textbf{147}_{\pm 44}$} & \cellcolor{codegray}\footnotesize{$\textbf{657}_{\pm 68}$} \\ 
\Xhline{1.0pt}
\end{tabular}
\end{center}}\end{minipage}
}
\vspace{3pt}
\subfloat[\qquad  \textbf{Left:} 
Learn old task in \textbf{\walkercolor{Walker}}~(\mono{stand}) \qquad \qquad \qquad \qquad \textbf{Right:}
Transfer to new task \textbf{\walkercolor{Walker}}~(\mono{walk})
]{
\centering
\begin{minipage}{0.49\linewidth}{\begin{center}
\footnotesize
\setlength{\tabcolsep}{5pt}
\renewcommand{\arraystretch}{1.0}
\begin{tabular}{l | c c |c }
\Xhline{1.0pt}
\multirow{2}{*}{Method} & \multicolumn{2}{c|}{   {Learn from scratch} } &  {Retest after} \\
&   {$100 k$} &   {$500 k$} &   {new task fine-tune} \\
\Xhline{1.0pt}
 {DrQ} &   \onepm{875}{76} & \onepm{973}{65}   &   \onepm{698}{57}  \\
 {Dreamer} &  \onepm{583}{21}&  \onebfpm{974}{31} &   \onepm{912}{19}\\
 {Resnet+SAC} &   \onepm{177}{32} &   \onepm{190}{24}&   \onepm{190}{24}\\
\cellcolor{codegray}{ {CtrlFormer}} & \cellcolor{codegray}   \onebfpm{877}{42}
& \cellcolor{codegray}   \onepm{954}{38}
& \cellcolor{codegray}  \onebfpm{950}{42}\\
\Xhline{1.0pt}
\end{tabular}
\end{center}}\end{minipage}
\begin{minipage}{0.49\linewidth}{\begin{center}
\footnotesize
\setlength{\tabcolsep}{5pt}
\renewcommand{\arraystretch}{1.0}
\begin{tabular}{l | c c | c c}
\Xhline{1.0pt}
\multirow{2}{*}{Method}  & \multicolumn{2}{c|}{ \footnotesize{Learn from scratch} } & \multicolumn{2}{c}{ \footnotesize{Learn with transfer} } \\
 & \footnotesize{$100 k$} & \footnotesize{$500 k$} & \footnotesize{$100 k$} & \footnotesize{$500 k$} \\
\Xhline{1.0pt}
\footnotesize{DrQ} & \footnotesize{$5 0 4_{\pm 191}$} & \footnotesize{$\mathbf{9 4 7}_{\pm 101}$} & \footnotesize{$321_{\pm 54}$} & \footnotesize{$947_{\pm 36}$} \\
\footnotesize{Dreamer} & \footnotesize{$277_{\pm 12}$} & \footnotesize{$897_{\pm 49}$} & \footnotesize{$851_{\pm 44}$} & \footnotesize{$949_{\pm 22}$} \\
\footnotesize{Resnet+SAC} & \footnotesize{$63_{\pm 7}$} & \footnotesize{$148_{\pm 12}$} & \footnotesize{$63_{\pm 7}$ }& \footnotesize{$148_{\pm 12}$} \\
\cellcolor{codegray}\footnotesize{CtrlFormer} & \cellcolor{codegray}\footnotesize{$\textbf{593}_{\pm 52}$} & \cellcolor{codegray}\footnotesize{$903_{\pm 43}$} & \cellcolor{codegray}\footnotesize{$\textbf{857}_{\pm 47}$} & \cellcolor{codegray}\footnotesize{$\mathbf{959}_{\pm 42}$} \\
\Xhline{1.0pt}
\end{tabular}
\end{center}}\end{minipage}
}
\\
\vspace{-15pt}
\caption{\textbf{
Transferring the state representation among tasks under same domain.} We list the sample efficiency of learning from scratch in both the previous task and the score retested after the new task fine-tuning in the sub-tables on the \textit{left side}, and the comparison between learning new task from scratch and learning new task with transfer in the sub-tables on the \textit{right side}.}\label{table:Transferring_tasks_in_same_domain}
\end{table*}

%% file: src/table2.tex
\begin{table*}[ht]
\centering
\subfloat[
Transfer from \textbf{\reachercolor{Reacher}}(\mono{easy}) to \textbf{\fingercolor{Finger}}(\mono{turn-easy})
]{
\centering
\begin{minipage}{0.49\linewidth}{\begin{center}
\footnotesize
\setlength{\tabcolsep}{1.2pt}
\renewcommand{\arraystretch}{1.0}
\begin{tabular}{l | c| cc |c}
\Xhline{1.0pt}
\multirow{2}{*}{Method}  & Scratch~\scriptsize{(previous)} & \multicolumn{2}{c|}{Transfer ~\scriptsize{(new task)}} & Retest~\scriptsize{(previous)} \\
 & 500 $k$ &  100 $k$ & 500 $k$ & 500 $k$ \\
\Xhline{1.0pt}
DrQ  &  \onebfpm{971}{27} & \onepm{ 283}{121} & \onepm{332}{96} & \onepm{124}{22}\\
Resnet+SAC & \onepm{382}{299} & \onepm{298}{17} & \onepm{300}{29} & \onepm{   382}{299} \\
\cellcolor{codegray} \model & \cellcolor{codegray}\onepm{918}{33}  & \cellcolor{codegray}\onebfpm{299}{38} & \cellcolor{codegray}\onebfpm{547}{56} & \cellcolor{codegray}\onebfpm{889}{34} \\
\Xhline{1.0pt}
\end{tabular}
\end{center}}\end{minipage}
}
\subfloat[
Transfer from  \textbf{\fingercolor{Finger}}(\mono{turn-easy}) to  \textbf{\reachercolor{Reacher}}(\mono{easy})
]{
\centering
\begin{minipage}{0.49\linewidth}{\begin{center}
\footnotesize
\setlength{\tabcolsep}{1.2pt}
\renewcommand{\arraystretch}{1.0}
\begin{tabular}{l | c| cc |c}
\Xhline{1.0pt}
\multirow{2}{*}{Method} & Scratch~\scriptsize{(previous )} & \multicolumn{2}{c|}{Transfer~\scriptsize{ (new task)}} & Retest~\scriptsize{(previous)} \\
 & 500 $k$ &  100 $k$ & 500 $k$ & 500 $k$ \\
\Xhline{1.0pt}
  {   {DrQ}} &   
\onebfpm{448}{65} & \onepm{203}{87}&   \onepm{693}{282} &  \onepm{184}{57}\\
  {   {Resnet+SAC}} &   {$300_{\pm 29}$} &   {$322_{\pm 285}$} &   {$382_{\pm 299}$} &   {$300_{\pm 29}$} \\
\cellcolor{codegray}  {   {CtrlFormer}} 
& \cellcolor{codegray}  \onepm{424}{35} 
& \cellcolor{codegray}  \onebfpm{416}{117} 
& \cellcolor{codegray} \onebfpm{770}{71}  
& \cellcolor{codegray} \onebfpm{409}{31} \\
\Xhline{1.0pt}
\end{tabular}
\end{center}}\end{minipage}
}
\vspace{-10pt}
\caption{\textbf{Transferring the state representation among tasks under cross-domain.} The first column is the performance of learning the previous task from scratch, the second column is the performance of learning the new task with the state representation transferring from the previous task, and the third column is the retest performance of the previous task using the latest encoder after learning the new task.}\label{table:cross_domain}
\end{table*}

%% file: src/6-conclusion.tex
\section{Conclusion}
In this paper, we propose a novel representation learning framework \model that learns a transferable state representation for visual control tasks via a sample-efficient vision transformer. \model explicitly learns the attention among the current task, the tasks it learned before, and the observations. Furthermore, each task is co-trained with contrastive learning as an auxiliary task to improve the sample efficiency when learning from scratch. Various experiments show that \model outperforms previous work in terms of transferability and the great potential to be extended in multiple sequential tasks. 
We hope our work could inspire rethinking the transferability of state representation learning for visual control and exploring the next generation of visual RL framework.

\noindent \textbf{Acknowledgement.} Ping Luo is supported by the General Research Fund of HK No.27208720 and 17212120.

%% file: src/7-appendix.tex
\appendix

\section{Extended Background}
\label{section:extended_background}

\paragraph{Soft Actor-Critic} 
The Soft Actor-Critic (SAC)~\cite{haarnoja2018sac} learns a state-action value function $Q_\theta$, a stochastic policy $\pi_\theta$  and a temperature $\alpha$ to find an optimal policy for an MDP $(S, A, p, r, \gamma)$ by optimizing a $\gamma$-discounted maximum-entropy objective~\cite{ziebart2008maxent}.
$\theta$ is used generically to denote the parameters updated through training in each part of the model. The actor policy $\pi_{\theta}(a_t|s_t)$ is a parametric $\text{tanh}$-Gaussian that given $s_t$ samples  $a_t = \text{tanh}(\mu_\theta(s_t)+ \sigma_\theta(s_t) \epsilon)$, where $\epsilon \sim N(0, 1)$ and $\mu_\theta$ and $\sigma_\theta$ are parametric mean and standard deviation.

The policy evaluation step  learns  the critic $Q_\theta(s_t, a_t)$ network by optimizing a single step of the soft Bellman residual
\begin{equation}
\begin{split}
    J_Q(\mathcal{D}) &= E_{\substack{( s_t,a_t, s'_t) \sim \mathcal{D} \\ a_t' \sim \pi(\cdot|s_t')}}[(Q_\theta(s_t, a_t) - y_t)^2] \\
    y_t &= r(s_t, a_t) + \gamma [Q_{\theta'}(s'_t, a'_t) - \alpha \log \pi_\theta(a'_t|s'_t)] , 
\end{split}  
\end{equation}
where $\mathcal{D}$ is a replay buffer of transitions, $\theta'$ is an exponential moving average of the weights. SAC uses clipped double-Q learning~\cite{hasselt2015doubledqn,fujimoto2018td3}, which we omit from our notation for simplicity but employ in practice.

The policy improvement step then fits the actor policy $\pi_\theta(a_t|s_t)$ network by optimizing the objective
\begin{align*}
    J_\pi(\mathcal{D}) &= E_{s_t \sim \mathcal{D}}[ D_{KL}(\pi_\theta(\cdot|s_t) || \exp\{\frac{1}{\alpha}Q_\theta(s_t, \cdot)\})].
\end{align*}
Finally, the temperature $\alpha$ is learned with the loss
\begin{align*}
    J_\alpha(\mathcal{D}) &= E_{\substack{s_t \sim \mathcal{D} \\ a_t \sim \pi_\theta(\cdot|s_t)}}[-\alpha \log \pi_\theta(a_t|s_t) - \alpha \mathcal{H}],
\end{align*}
where $\mathcal{H}$ is the target entropy hyper-parameter that the policy tries to match.

\paragraph{Deep Q-learning} DQN~\cite{mnih2013dqn} also learns a convolutional neural net to approximate Q-function over states and actions. The main difference is that DQN operates on discrete actions spaces, thus the policy can be directly inferred from Q-values. The parameters of DQN are updated by optimizing the squared residual error
\begin{align*}
    J_Q(\mathcal{D}) &= E_{( s_t,a_t, s'_t) \sim \mathcal{D}}[(Q_\theta(s_t, a_t) - y_t)^2] \\
    y_t &= r(s_t, a_t) + \gamma \max_{a'} Q_{\theta'}(s'_t, a') . \\
\end{align*}
In practice, the standard version of DQN is frequently combined with a set of tricks that improve performance and training stability, wildly known as Rainbow~\cite{hasselt2015doubledqn}.

\section{Supplumentary Materials of Experiments}\label{sec:settings}
\label{section:hyperparams}

Our PyTorch code is implanted based on the Timm~\cite{rw2019timm},  Pytorch version  SAC~\cite{haarnoja2018sac}, and the official code of DrQ-v1\cite{kostrikov2020image}. All the experiments are run on the GeForce RTX 3090 with 5 seeds.
\subsection{Detailed Structure of Vision Transformer in \model}
\begin{figure}[htbp]
    \centering
    \includegraphics[width=0.85\linewidth]{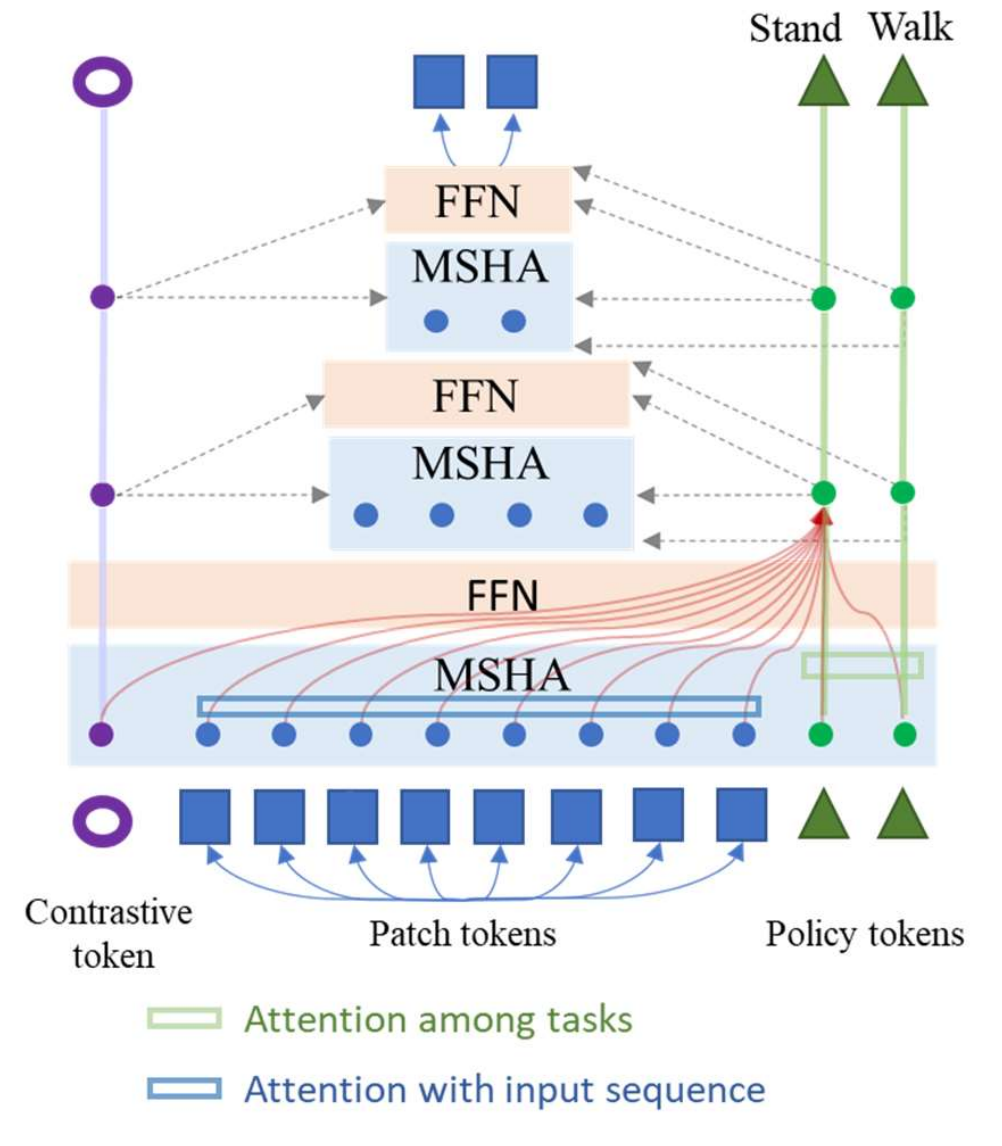}
    \caption{The detailed structure of \model}
    \label{fig:detailed-structure}
    \vspace{-10pt}
\end{figure}

We utilize a simple pyramidal vision transformer as the encoder of \model, which has 3 stages and each stage  contains 3 blocks. With the 192 dimension output of the vision transformer, we add a fully-connected layer to map the feature dimension to 50, and apply \text{tanh} nonlinearity to the $50$ dimensional output as the final output of the encoder. 
The  detailed structure of \model is shown as Figure \ref{fig:detailed-structure}
The  hyper-parameters are listed in Table \ref{tab:Hyper-parameter-former}. The implantation is based on Timm ~\cite{rw2019timm}, which is a collection of SOTA computer vision models with the ability to reproduce ImageNet training results.
We train the transformer model using Adamw~\citep{kingma2014adam} as optimizer with $\beta_1=0.9$, $\beta_2=0.999$, a batch size of 512 and apply a high weight decay of $0.1$.
\begin{table}[b]
    \centering
\begin{tabular}{l|c}
\hline
 Parameters& Value\\
\hline
 Image size & $84 \times 84$ \\
 Patch size & 8 \\
 Num patches & 196 \\
 Input channels & 9 \\
 Embedding dim & 192 \\
 Depth & 9 \\
 Num stage & 3 \\
 Num blocks per stage & 3 \\
 num heads & 3 \\
\hline
\end{tabular}
\caption{Hyper-parameter of \model}
\label{tab:Hyper-parameter-former}
\end{table}
\subsection{Actor and Critic Networks}
\label{app: detailed-structure}
The structure of actor and critic networks are the same in \model and DrQ(CNN+multiple heads). Clipped double Q-learning~\cite{hasselt2015doubledqn, fujimoto2018td3} is used for the critic, where each $Q$-function is parametrized as a 3-layer MLP with ReLU activations after each layer except for the last. The actor is also a 3-layer MLP with ReLUs that outputs mean and covariance for the diagonal Gaussian that represents the policy. The hidden dimension is set to $1024$ for both the critic and actor. 

\subsection{CNN-based Encoder Network used in DrQ} 
The CNN-based model ~\cite{kostrikov2020image} employs an encoder consists of four convolutional layers with $3\times 3$ kernels and $32$ channels, which is same to DrQ~\cite{kostrikov2020image}. The ReLU activation is applied after each convolutional layer. We use stride to $1$ everywhere, except for the first convolutional layer, which has stride $2$. The output of the convolutional network is fed into a single fully-connected layer normalized by LayerNorm~\cite{ba2016layer}. Finally, we apply \text{tanh} non-linearity to the $50$ dimensional output of the fully-connected layer. We initialize the weight matrix of fully-connected and convolutional layers with the orthogonal initialization~\cite{saxe2013ortho}.

The actor and critic networks share the weights of the convolutional layers of the encoder. Furthermore, only the critic optimizer is allowed to update these weights (e.g. the gradients from the actor is stopped before they propagate to the shared CNN layers).

\subsection{Detailed Training and Evaluation Setup for Visual Control Tasks}
\label{app:Detailed_Training}
\begin{figure}
    \centering
   \includegraphics[width=0.55\textwidth]{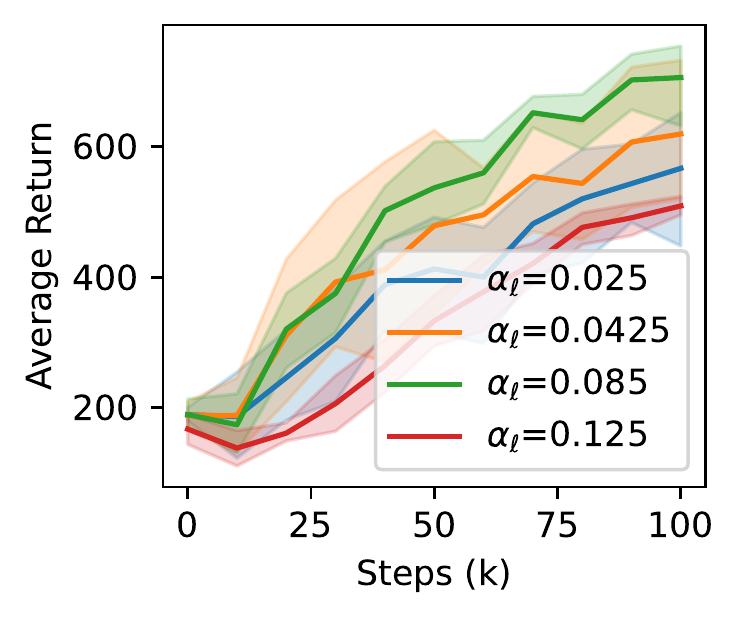}
   \vspace{-20pt}
    \caption{Training curves with different $\alpha_{\ell}$}
    \label{fig:Training curves}
\end{figure}

The agent first collects $1000$ seed observations using a random policy. The further training observations are collected by sampling actions from the current policy. The agent performs one training update every time when receiving a new observation. 
The action repeat parameters are used as same as the DrQ~\cite{kostrikov2020image}, and is listed in Table~\ref{table:action_repeat}, and the number of training observations is only a fraction of the environment steps (e.g. a $1000$ steps episode at action repeat $8$ will only result in $125$ training observations). 
In order to avoid damaging the \model due to the low-quality policy gradient caused by the inaccurate policy network and Q-networks at the beginning of the behavior learning in the new task, 
the learning rate of \model is set $\alpha_{l}$ times lower than  policy network and Q-networks.  In this way, the agent is guided to quickly learn the policy of the current task based on the previously learned representation and fine-tune the model according to the relationship between the current task, the previous tasks, and the input. The training curves using different $\alpha_{\ell}$  are shown in the Figure \ref{fig:Training curves}, showing that $\alpha_{\ell}$ in the range of 0.04-0.08 is appropriate. 
We evaluate our agent every $10000$ environment step by computing the average episode return over $10$ evaluation episodes as same as DrQ~\cite{kostrikov2020image}. During the evaluation, we take the mean of the policy output action instead of stochastic sampling. In Table~\ref{table:hyper_params} we provide a comprehensive overview of all the other hyper-parameters.
\begin{table}[ht!]
\centering
\begin{tabular}{l|c}
\hline
Task name        & Action repeat \\
\hline
Cartpole Swingup &  $8$ \\
Cartpole Swingup sparse &  $8$ \\
Cartpole balance &  $8$ \\
Cartpole balance sparse &  $8$ \\
Reacher Easy &  $4$ \\
Reacher Hard &  $4$ \\
Finger Turn easy & $2$ \\
Finger Turn Hard & $2$ \\
Walker Walk & $2$ \\
Walker stand & $2$ \\
\hline
\end{tabular}
\caption{\label{table:action_repeat} The action repeat hyper-parameter used for each task.}
\end{table}

\subsection{Image Preprocessing and Augmentation}
\label{augmentaion}
We construct an observational input as a $3$-stack of consecutive frames, which is the same as DQN~\cite{mnih2013dqn} and DrQ~\cite{kostrikov2020image}, where each frame is an RGB rendering of size $ 84 \times 84$ from the camera. We then divide each pixel by $255$ to scale it down to $[0, 1]$ range as the input of the encoder. 

The images from the DeepMind control suite are $84\times 84$. 
The image augmentation used in \ref{subsec: RL} and \ref{sec:Contrastive} is implanted by random crop which is also used in DrQ ~\cite{kostrikov2020image}. We pad each side by $4$ repeating boundary pixels and then select a random $84\times 84$ crop, yielding the original image shifted by $\pm 4$ pixels. This procedure is repeated every time an image is sampled from the replay buffer.
\begin{table}[hb!]
\centering
\begin{tabular}{l|c}
\hline
Parameter        & Setting \\
\hline
Replay buffer capacity & $100000$ \\
Seed steps & $1000$ \\
Batch size & $512$ \\
Discount $\gamma$ & $0.99$ \\
Optimizer & Adamw \\
Learning rate & $10^{-4}$ \\
Critic target update frequency & $2$ \\
Critic Q-function soft-update rate $\tau$ & $0.01$ \\
Actor update frequency & $2$ \\
Actor log stddev bounds & $[-10, 2]$ \\
Init temperature & $0.1$ \\

\hline
\end{tabular}\\
\caption{\label{table:hyper_params} Hyper-parameters of downstream reinforcement learning.}
\end{table}

\subsection{Implementation Details of Contrastive Learning}
\label{sec:impl_detail_contrastive}
We implanted the contrastive optimization by  BYOL ~\cite{grill2020bootstrap}, which is a typical approach to self-supervised image representation learning. 

BYOL relies on two neural networks, referred to as \textit{online} and \textit{target} networks, that interact and learn from each other and do not rely on negative pairs. 
From an augmented view of an image, we train the online network $f_\netparams$ to predict the target network $f_\targetparams$ representation of the same image under a different augmented view. 
At the same time, we update the target network with a slow-moving average of the online network. 

The output of the contrastive token in \model is a 192 dimension vector. We project it to a 96 dimension vector by a multi-layer perceptron (MLP) and similarly for the target projection.
This MLP consists of a linear layer with output size 384 followed by batch normalization~\cite{ioffe2015batch}, rectified linear units (ReLU)~\cite{agarap2018deep}, and a final linear layer with output dimension 96. The predictor $q_{\netparams}$ uses the same architecture as the projector.
For the target network, the exponential moving average parameter $\tau$ starts from $\tau_\text{base} = 0.996$ and is increased to one during training.  

\subsection{Additional Results}

\textbf{Ablation on Q-regularization}.
To illustrate the effect of the Q-regularization technique, we compare the transferability of \model,  \model without Q-regularization, and DrQ without Q-regularization.
Table \ref{tab:rq_ab} shows that the  Q-regularization technique helps to improve the performance of CtrlFormer, and \model still outperforms DrQ on transferability without Q-regularization. 
\begin{table}[h]
\caption{Ablation on Q-regularization by transferring from Reacher(easy) to Reacher(hard).
}
\scriptsize
\label{tab:rq_ab}
\setlength{\tabcolsep}{0.5pt}
\begin{tabular}{l|c|c|c|c|c}
\hline & Scratch Task1& Retest & Scratch Task2  & Transfer& Benifit \\
\hline Our w/ rQ & $973_{\pm 53}$ & $906_{\pm 31}$ & $548_{\pm 124}$ & $657_{\pm 68}$ &{${+16.5 \%}$}  \\
\hline Our w/o rQ & $774_{\pm 32}$ & $738_{\pm 54}$ & $474_{\pm 56}$ & $551_{\pm 57}$ & {${+13.8 \%}$ }\\
\hline DRQ w/o rQ & $756_{\pm 47}$ & $329_{\pm 58}$ & $481_{\pm 198}$ & $410_{\pm 47}$ & {${-14.76 \%}$} \\
\hline 
\end{tabular}
\vspace{-10pt}
\end{table}

{\textbf{Contrastive co-training on other baselines.}} 
We provide an additional baseline that applies contrastive co-training on the CNN-based model and compare it with CtrlFormer. The results show DrQ with contrastive co-training (DrQ-C) achieves better transferring results from Walker-Stand (T1) to Walker-Walk (T2) than the original DrQ algorithm. But it still suffers from catastrophic forgetting.
\begin{table}[h]
\footnotesize
\setlength{\tabcolsep}{1pt}
\begin{tabular}{l|c|c}
\hline & Retest(T1 500k) & Transfer T2 (100k) \\
\hline DrQ&$698_{\pm 57}$ & $321_{\pm 54}$ \\
\hline DrQ-C& $707_{\pm 68}$ & $472_{\pm 68}$ \\
\hline Our & $\textbf{950}_{\pm 42}$ & $\textbf{857}_{\pm 47}$ \\
\hline
\end{tabular}
\vspace{-10pt}
\caption{Ablation on Contrastive co-training}
\end{table}

\subsection{PyTorch-style Pseudo-code}
\label{app:Pseudo-code}
We provide detailed PyTorch-style pseudo-codes of the method we visualize the policy attention shown in Figure \ref{fig:visualization}, and the method we update the encoder, the actor and the critic. The pseudo-code of Grad-CAM is shown in Listing 1. The pseudo-code of the actor and the entropy temperature is shown in Listing 2. The pseudo-code of the encoder and the critic is shown in Listing3.

\subsection{DMControl Benchmark}
\label{app:dmc_description}
The DeepMind Control Suite (DMControl) ~\cite{tassa2018deepmind} is a set of stable, well-tested continuous control tasks that are easy to use and modify. DMControl contains many well designed tasks, which are written in \href{https://www.python.org/}{Python} and physical models are defined using \href{http://mujoco.org/book/modeling.html}{MJCF}. It is currently one of the most recognized standard test environments for visual control.
The domain in DMControl refers to a physical model, while a task refers to an instance of that model with a particular MDP structure. For example, the difference between the \mono{swingup} and \mono{balance} tasks of the \mono{cartpole} domain is whether the pole is initialized pointing downwards or upwards, respectively.
We list the detailed descriptions of the domains used in this paper below, names are followed by three integers specifying the dimensions of the state, control and observation spaces i.e.\ $\Bigl(\dim(\mathcal{S}),\dim(\mathcal{A}),\dim(\mathcal{O})\Bigr)$. 

\myfigure{
\textbf{\walkercolor{Walker} (18, 6, 24):} 
The \textbf{\walkercolor{Walker}} domain contains a series of control tasks for two-legged robots. 
In the \mono{stand} task, the reward is a combination of terms encouraging an upright torso and some minimal torso height. The \mono{walk} and \mono{run} tasks include a component encouraging forward velocity.
}{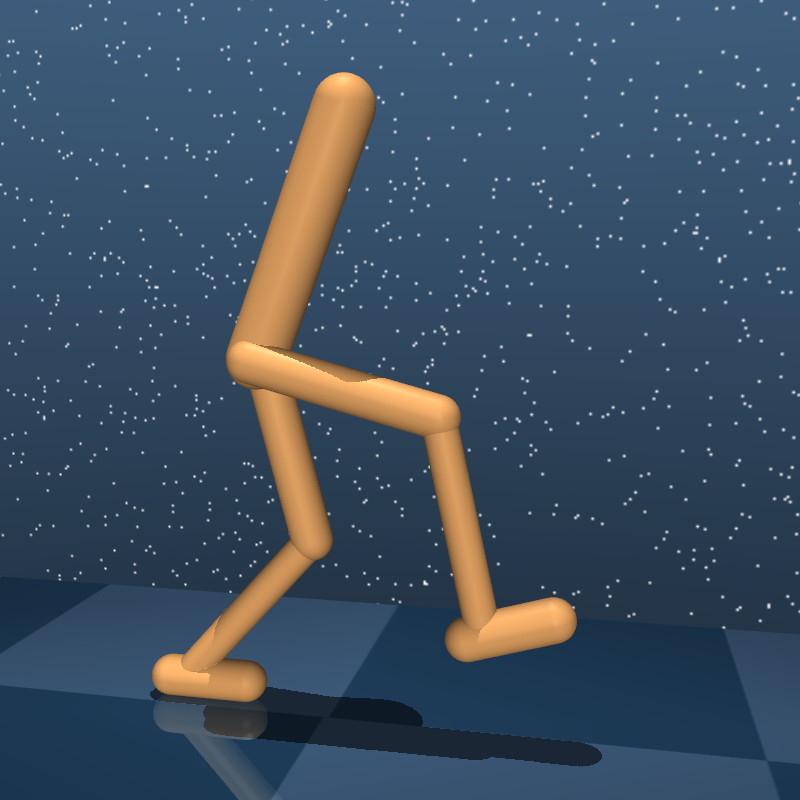}
\vspace{-10pt}

\myfigure{
\textbf{\fingercolor{Finger}(6, 2, 12):}  \textbf{ \fingercolor{Finger}} domain aims to rotate a body on an unactuated hinge. In the \mono{turn\_easy} and \mono{turn\_hard} tasks, the tip of the free body must overlap with a target (the target is smaller for the \mono{turn\_hard} task). In the \mono{spin} task, the body must be continually rotated.
}{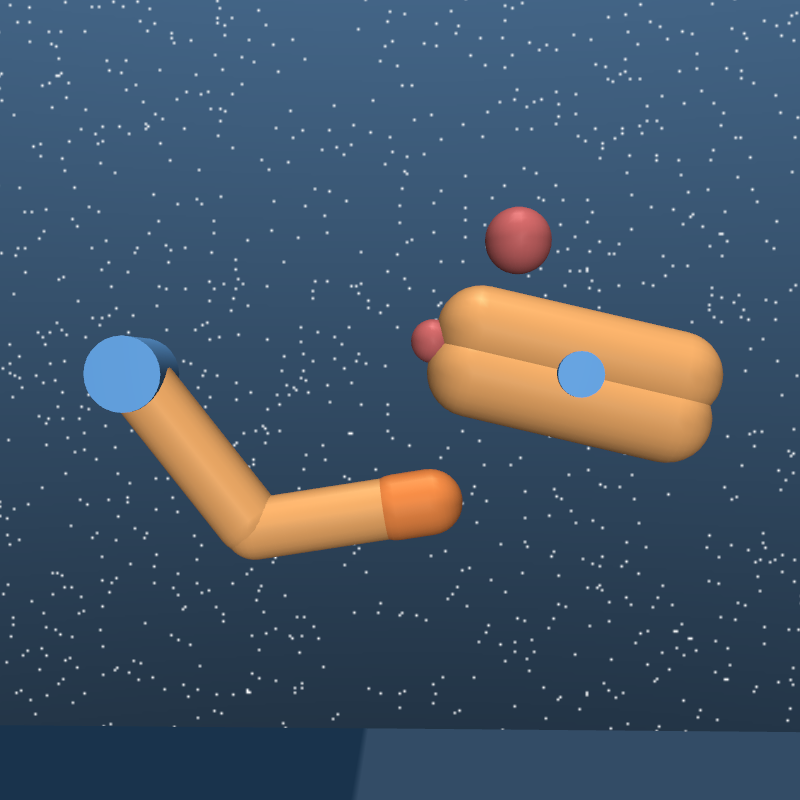}
\vspace{-10pt}

\myfigure{
\textbf{\cartpolecolor{Cartpole}(4, 1, 5):}
\textbf{\cartpolecolor{Cartpole}} domain aims to control the pole attached by an un-actuated joint to a cart. 
In both the \mono{swingup} task and the \mono{swingup-sparse} task,
the pole starts pointing down and aim to make the unactuated pole keeping upright upward by applying forces to the cart, while in \mono{balance} and \mono{balance\_sparse} the pole starts near the upright. }{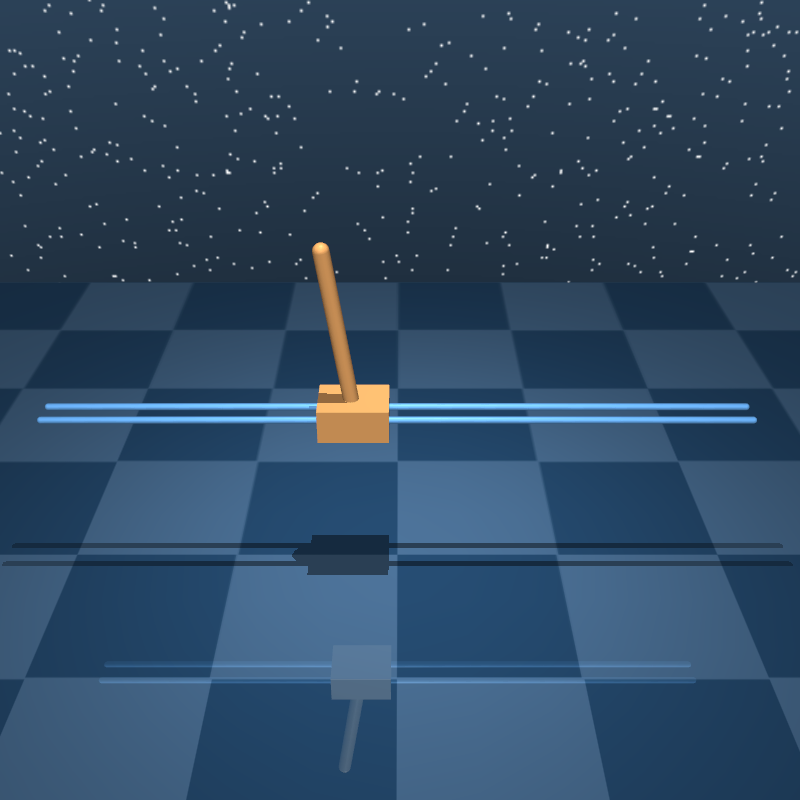}
\vspace{-10pt}

\myfigure{
\textbf{\reachercolor{Reacher} (4, 2, 7):} \textbf{\reachercolor{Reacher}} domain aims to control the two-link planar reach a randomised target location. The reward is one when the end effector penetrates the target sphere. In the \mono{easy} task the target sphere is bigger than on the \mono{hard} task (shown on the left).  
}{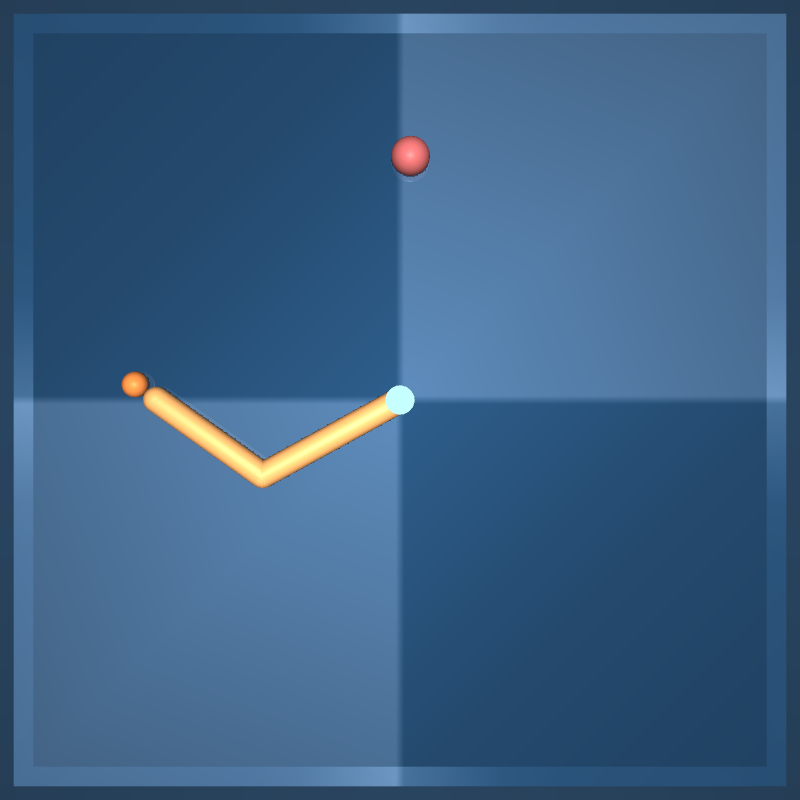}

\section{Visualization}
\label{app:Visual}
\begin{figure*}[ht]
    \centering
    \begin{subfigure}{0.24\linewidth}
        \centering
        \includegraphics[width=\linewidth]{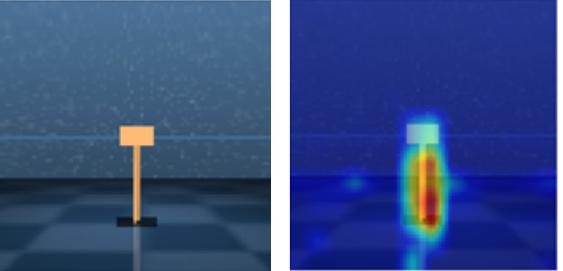}
        \caption{Cartpole-swingup}
    \end{subfigure}%
    \hfill
    \begin{subfigure}{0.24\linewidth}
        \centering
        \includegraphics[width=\linewidth]{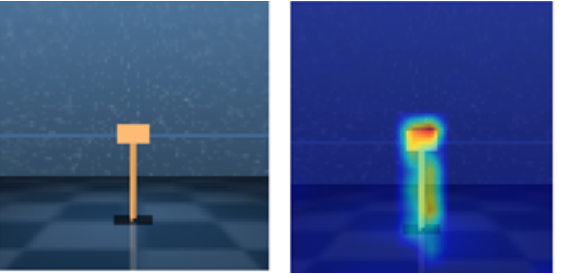}
        \caption{cartpole-swingup-sparse}
    \end{subfigure}
    \hfill
    \begin{subfigure}{0.24\linewidth}
        \centering
        \includegraphics[width=\linewidth]{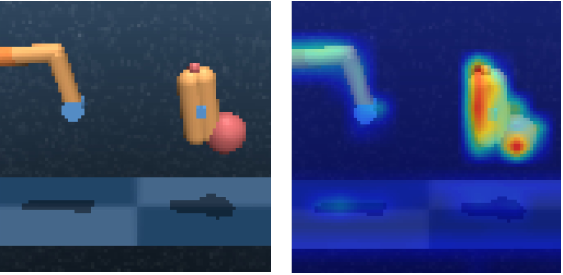}
        \caption{Finger-turn-easy}
    \end{subfigure}%
    \hfill
    \begin{subfigure}{0.24\linewidth}
        \centering
        \includegraphics[width=\linewidth]{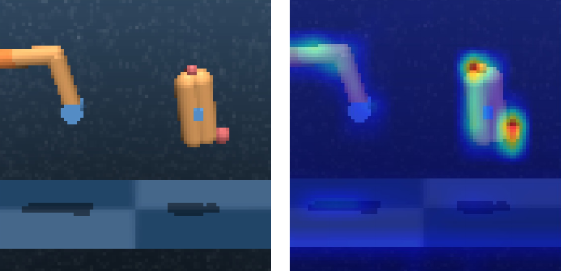}
        \caption{Finger-turn-hard}
    \end{subfigure}%
    \hfill
    \begin{subfigure}{0.24\linewidth}
        \centering
        \includegraphics[width=\linewidth]{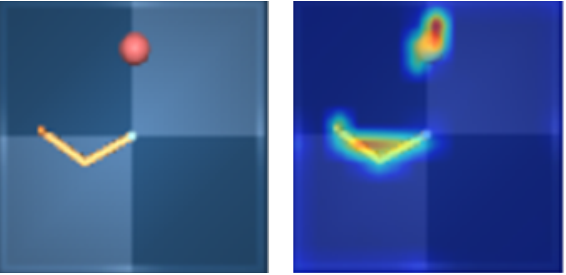}
        \caption{Reacher-easy}
    \end{subfigure}%
    \hfill
    \begin{subfigure}{0.24\linewidth}
        \centering
        \includegraphics[width=\linewidth]{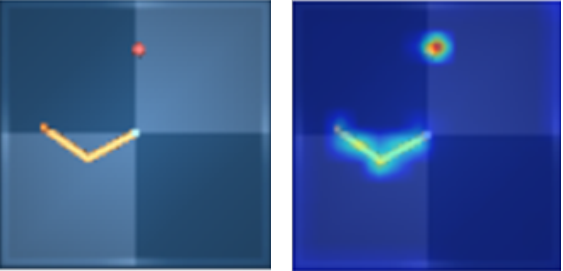}
        \caption{Reacher-hard}
    \end{subfigure}%
    \hfill
    \begin{subfigure}{0.24\linewidth}
        \centering
        \includegraphics[width=\linewidth]{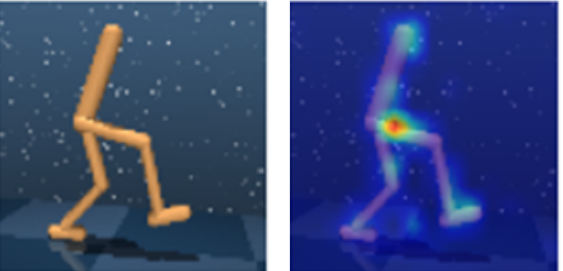}
        \caption{Walker-stand}
    \end{subfigure}%
    \hfill
    \begin{subfigure}{0.24\linewidth}
        \centering
        \includegraphics[width=\linewidth]{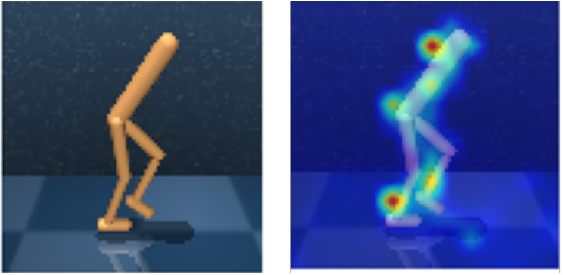}
        \caption{Walker-walk}
    \end{subfigure}%
    \hfill
    \caption{Visualization of the \model's attention on the input image}
    \label{fig:att_vis_ours}
\end{figure*}

\begin{figure*}[ht]
    \centering
    \begin{subfigure}{0.24\linewidth}
        \centering
        \includegraphics[width=\linewidth]{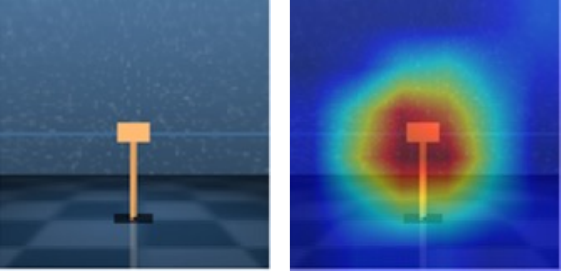}
        \caption{Cartpole-swingup}
    \end{subfigure}%
    \hfill
    \begin{subfigure}{0.24\linewidth}
        \centering
        \includegraphics[width=\linewidth]{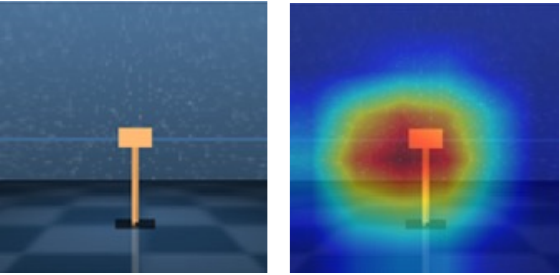}
        \caption{cartpole-swingup-sparse}
    \end{subfigure}
    \hfill
    \begin{subfigure}{0.24\linewidth}
        \centering
        \includegraphics[width=\linewidth]{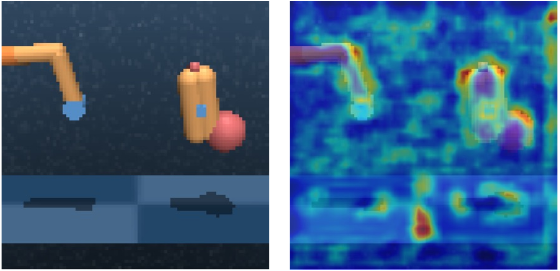}
        \caption{Finger-turn-easy}
    \end{subfigure}%
    \hfill
    \begin{subfigure}{0.24\linewidth}
        \centering
        \includegraphics[width=\linewidth]{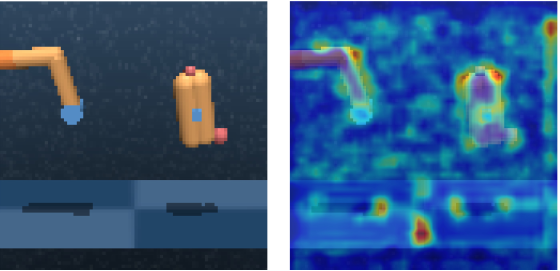}
        \caption{Finger-turn-hard}
    \end{subfigure}%
    \hfill
    \begin{subfigure}{0.24\linewidth}
        \centering
        \includegraphics[width=\linewidth]{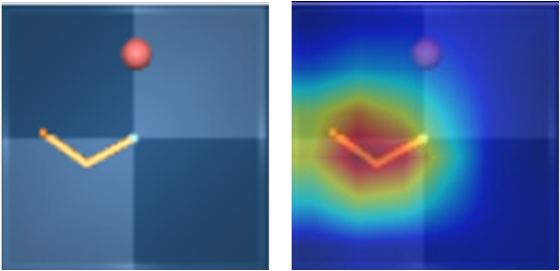}
        \caption{Reacher-easy}
    \end{subfigure}%
    \hfill
    \begin{subfigure}{0.24\linewidth}
        \centering
        \includegraphics[width=\linewidth]{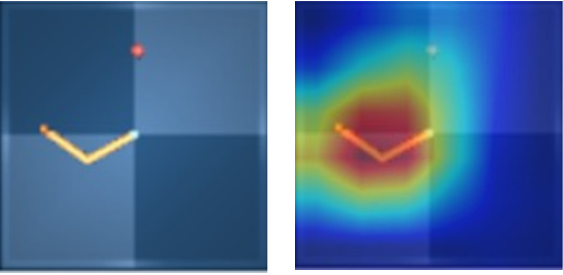}
        \caption{Reacher-hard}
    \end{subfigure}%
    \hfill
    \begin{subfigure}{0.24\linewidth}
        \centering
        \includegraphics[width=\linewidth]{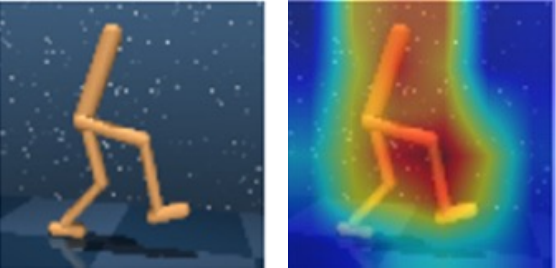}
        \caption{Walker-stand}
    \end{subfigure}%
    \hfill
    \begin{subfigure}{0.24\linewidth}
        \centering
        \includegraphics[width=\linewidth]{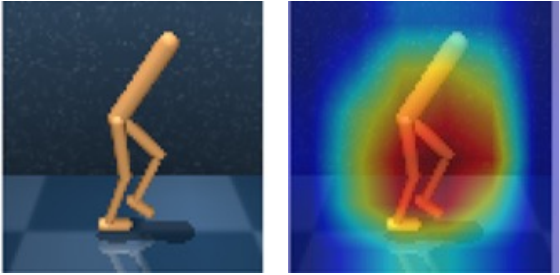}
        \caption{Walker-walk}
    \end{subfigure}%
    \hfill
    \caption{Visualization of the Resnet's attention on the input image}
    \label{fig:att_vis_resnet}
\end{figure*}

 We use the Grad-CAM ~\cite{selvaraju2016grad} method to visualize the encoder's attention on the input image, which is a typical technique for visualizing the regions of input that are "important". Grad-CAM uses the gradient information flowing to produce a coarse localization map of the important regions in the image.
From Figure~\ref{fig:att_vis_ours} and Figure~\ref{fig:att_vis_resnet} we can observe that our attention map is more focused on objects and task-relevant body parts, while the attention of the pre-trained ResNet-50 (same as the network used in the experiments) is disturbed by irrelevant information and not focused. In this way, our \model learns better policies by the representation highly relevant to the task.
The attention learned from similar tasks has similarities, but each has its own emphasis.

\clearpage
\onecolumn
\clearpage
\input{src/alg}

%% file: src/alg.tex
\definecolor{codegreen}{rgb}{0,0.8,0}
\definecolor{codegray}{rgb}{0.5,0.5,0.5}
\definecolor{codepurple}{rgb}{0.58,0,0.82}
\definecolor{backcolour}{rgb}{0.99,0.99,0.97}
\begin{lstlisting}[
float=htbp,
language=Python,
floatplacement=htbp,
xleftmargin=2em,
frame=single,
framexleftmargin=1.5em,
backgroundcolor=\color{backcolour},
belowskip=-1\baselineskip,
commentstyle=\color{codegreen},
keywordstyle=\color{magenta},
numberstyle=\tiny\color{codegray},
stringstyle=\color{codepurple},
backgroundcolor=\color{backcolour},
commentstyle=\color{codegreen},
basicstyle=\ttfamily\scriptsize,
breakatwhitespace=false,         
numbers=left,                    
breaklines=true,                 
captionpos=b,                    
keepspaces=true,                 
numbersep=5pt,                  
showspaces=false,                
showstringspaces=false,
showtabs=false,                  
tabsize=2,
label={lst:code1},
caption=PyTorch-style pseudo-code for attention visualization use in this paper.]
class ViTAttentionGradRollout:
    def __init__(self, model, attention_layer_name='attn_drop',
        discard_ratio=0.9):
        self.model = model
        self.discard_ratio = discard_ratio
        for name, module in self.model.named_modules():
            if attention_layer_name in name:
                module.register_forward_hook(self.get_attention)
                module.register_backward_hook(self.get_attention_gradient)

        self.attentions = []
        self.attention_gradients = []

    def get_attention(self, module, input, output):
        self.attentions.append(output.cpu())

    def get_attention_gradient(self, module, grad_input, grad_output):
        self.attention_gradients.append(grad_input[0].cpu())

    def __call__(self, input_tensor):
        self.model.zero_grad()
        output = self.model(input_tensor,detach=False)
        loss = (output).sum()
        loss.backward()
        return grad_rollout(self.attentions, self.attention_gradients,
            self.discard_ratio)
        \end{lstlisting}

\begin{lstlisting}[
float=htbp,
language=Python,
floatplacement=htbp,
xleftmargin=2em,
frame=single,
framexleftmargin=1.5em,
backgroundcolor=\color{backcolour},
belowskip=-1\baselineskip,
commentstyle=\color{codegreen},
keywordstyle=\color{magenta},
numberstyle=\tiny\color{codegray},
stringstyle=\color{codepurple},
backgroundcolor=\color{backcolour},
commentstyle=\color{codegreen},
basicstyle=\ttfamily\scriptsize,
breakatwhitespace=false,         
numbers=left,                    
breaklines=true,                 
captionpos=b,                    
keepspaces=true,                 
numbersep=5pt,                  
showspaces=false,                
showstringspaces=false,
showtabs=false,                  
tabsize=2,
label={lst:code2},
caption=PyTorch-style pseudo-code for the actor updating in down stream reinforcement learning.]
def update_actor_and_alpha(obs):
        # detach encoder layers
        dist = actor(obs, detach_encoder=True)
        action = dist.rsample()
        log_prob = dist.log_prob(action).sum(-1, keepdim=True)
        # detach encoder layes
        actor_Q1, actor_Q2 = critic(obs, action, detach_encoder=True)
        actor_Q = torch.min(actor_Q1, actor_Q2)
        actor_loss = (alpha.detach() * log_prob - actor_Q).mean()
        # optimize the actor
        actor_optimizer.zero_grad()
        actor_loss.backward()
        actor_optimizer.step()
        actor.log(logger, step)
        log_alpha_optimizer.zero_grad()
        alpha_loss = (alpha *
                     (-log_prob - target_entropy).detach()).mean()
        alpha_loss.backward()
        log_alpha_optimizer.step()
\end{lstlisting}
\clearpage
\begin{lstlisting}[
float=htbp,
language=Python,
floatplacement=htbp,
xleftmargin=2em,
frame=single,
framexleftmargin=1.5em,
backgroundcolor=\color{backcolour},
belowskip=-1\baselineskip,
commentstyle=\color{codegreen},
keywordstyle=\color{magenta},
numberstyle=\tiny\color{codegray},
stringstyle=\color{codepurple},
backgroundcolor=\color{backcolour},
commentstyle=\color{codegreen},
basicstyle=\ttfamily\scriptsize,
breakatwhitespace=false,         
numbers=left,                    
breaklines=true,                 
captionpos=b,                    
keepspaces=true,                 
numbersep=5pt,                  
showspaces=false,                
showstringspaces=false,
showtabs=false,                  
tabsize=2,
label={lst:code3},
caption=PyTorch-style pseudo-code for the updating of encoder and critic]
def update_critic_and_encoder(self, obs, obs_aug, action, reward, next_obs, next_obs_aug, not_done, logger, step):
        cons  = encoder.forward_rec(imgs)
        cons = encoder.byol_project(cons).detach()
        aug_cons = encoder.forward_cons(aug_imgs)
        project_cons = encoder.project(aug_cons)
        predict_cons = encoder.predict(project_cons)
        cons_loss = F.mse_loss(predict_cons,cons)
        with torch.no_grad():
            dist = actor(next_obs)
            next_action = dist.rsample()
            log_prob = dist.log_prob(next_action).sum(-1)
            target_Q1, target_Q2 = critic_target(next_obs,next_action)
            target_V = torch.min(target_Q1,
                                 target_Q2) - alpha.detach() * log_prob
            target_Q = reward + (not_done * discount * target_V)

            dist_aug = actor(next_obs_aug)
            next_action_aug = dist_aug.rsample()
            log_prob_aug = dist_aug.log_prob(next_action_aug).sum(-1)
            target_Q1, target_Q2 = critic_target(next_obs_aug, next_action_aug)
            target_V = torch.min(
                target_Q1, target_Q2) - alpha.detach() * log_prob_aug
            target_Q_aug = reward + (not_done * discount * target_V)
            target_Q = (target_Q + target_Q_aug) / 2
        # get current Q estimates
        current_Q1, current_Q2 = critic(obs, action)
        critic_loss = F.mse_loss(current_Q1, target_Q) + F.mse_loss(
            current_Q2, target_Q)
        Q1_aug, Q2_aug = critic(obs_aug, action)
        critic_loss += F.mse_loss(Q1_aug, target_Q) + F.mse_loss(
            Q2_aug, target_Q) + rec_loss
        logger.log('train_critic/loss', critic_loss, step)
        # Optimize the critic
        critic_optimizer.zero_grad()
        critic_loss.backward()
        critic_optimizer.step()
        critic.log(logger, step)
        \end{lstlisting}